%% file: main.tex
\pdfoutput=1
\documentclass[11pt, a4paper]{stanford}

\usepackage[authoryear, sort&compress, round]{natbib}
\usepackage{microtype}
\usepackage{subfiles}

\input{math_commands.tex}

\usepackage{hyperref}
\usepackage{url}
\usepackage{booktabs}
\usepackage{graphicx}

\usepackage{amsmath}
\usepackage[most]{tcolorbox}
\usepackage[utf8]{inputenc}
\usepackage{pifont}
\usepackage{fontawesome5}
\usepackage{xspace}
\usepackage[ruled, vlined, linesnumbered]{algorithm2e}
\usepackage{enumitem}
\usepackage{multirow}
\usepackage{xcolor}
\usepackage{wrapfig}
\usepackage{multirow}

\usepackage{tikz}
\usetikzlibrary{calc}
\usepackage{booktabs}

\newcommand{\licensepill}[1]{%
  \colorbox{gray!15}{\scriptsize\texttt{#1}}%
}

\definecolor{accentcolor}{HTML}{175E54}
\hypersetup{colorlinks=true, citecolor=accentcolor, linkcolor=accentcolor, urlcolor=accentcolor}

\title{\dataset: Why Coding Agents Cannot be Your Teammates Yet}

\author{%
\bfseries\fontsize{9}{13}\selectfont%
Arpandeep Khatua$^{1*}$, Hao Zhu$^{1*}$, \par Peter Tran$^{2**}$, Arya Prabhudesai$^{2**}$, Frederic Sadrieh$^{2**}$,
Johann K. Lieberwirth$^{2**}$,\par Xinkai Yu$^{1}$, Yicheng Fu$^{1}$, Michael J. Ryan$^{1}$, Jiaxin
Pei$^{1}$, Diyi Yang$^{1}$ \par
\vspace{0.5em}
$^{1}$Stanford University \qquad $^{2}$SAP Labs US \qquad $^{*}$Equal Contribution \qquad $^{**}$Equal Contribution \qquad \texttt{https://cooperbench.com}
}

\newcommand{\dataset}{CooperBench\xspace}

\makeatletter
\renewcommand{\maketitle}{\bgroup
\setlength{\parindent}{0pt}
\begin{adjustwidth}
  {0pt}{24pt}
  \begin{flushleft}
    { {\raggedright \titlefont \@title\par}%
    \vskip11pt {\raggedright \@author\par} \vskip20pt%
    }%
  \end{flushleft}
\end{adjustwidth}
\egroup

\begin{figure}[!h]
  \centering
  \makebox[\textwidth][c]{%
  \includegraphics[width=1\linewidth]{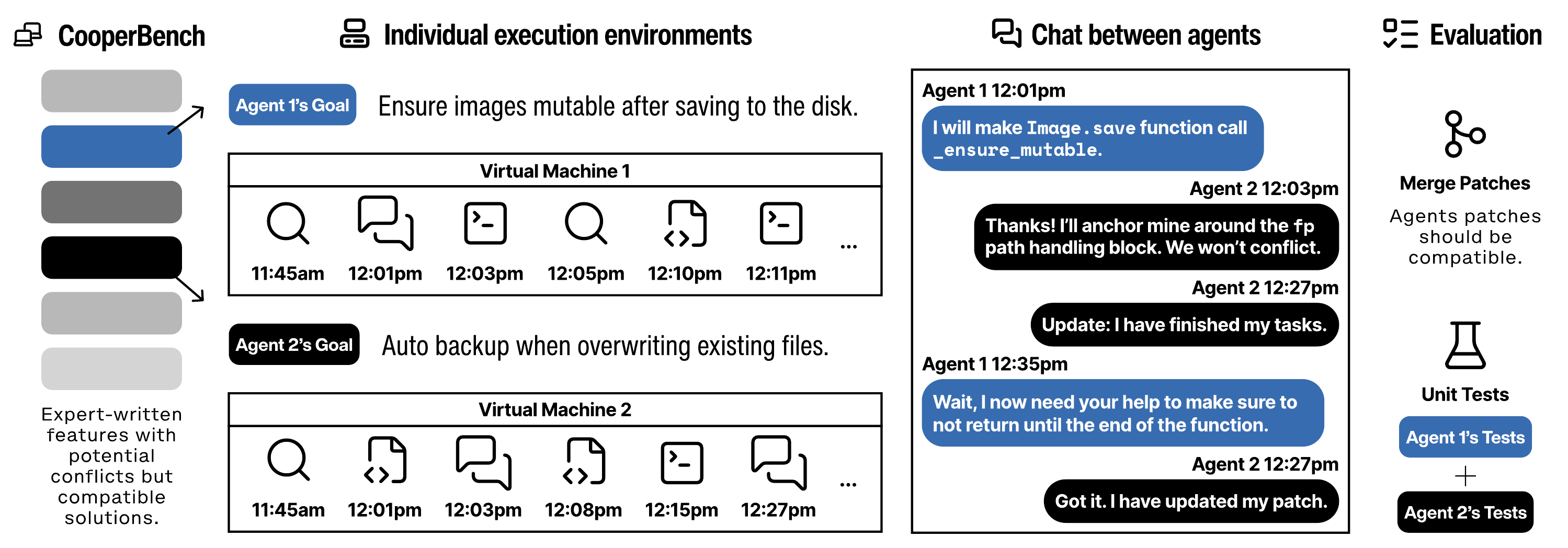}%
  }
  \caption{\small The \dataset 
  benchmark draws tasks for two agents from a pool of features with
  potential conflicts. The agents execute the tasks in their individual environments, communicating
  in real time to coordinate. Success is measured by whether the resulting code changes by both agents
  are compatible and pass the requirements for both features.}
  \label{fig:teaser}
\end{figure}

{%
{\abscontent} }%
\thispagestyle{firststyle}
}%
\makeatother

\begin{abstract}
  Resolving team conflicts requires not only task-specific competence, but also social intelligence to find common ground and build consensus. 
 Similarly, as AI agents increasingly collaborate on complex work, they must develop coordination capabilities to function as effective teammates. Yet we hypothesize that current agents lack these capabilities.
  To test this hypothesis, we introduce \textbf{\dataset}, a benchmark of over 600 collaborative coding
  tasks across 12 libraries in 4 programming languages. Each task assigns two agents different
  features that can be implemented independently but may conflict without proper coordination. Tasks are
  grounded in real open-source repositories with expert-written tests. Evaluating state-of-the-art
  coding agents, we observe \textbf{the curse of coordination}: agents achieve on average \textbf{30\% lower
  success rates} when working together compared to performing both tasks individually, across the full
  spectrum of task difficulties. This contrasts sharply with human teams, where adding teammates typically
  improves rather than diminishes productivity. Our analysis reveals three key issues: (1) communication
  channels become jammed with vague, ill-timed, and inaccurate messages; (2) even with effective
  communication, agents deviate from their commitments; and (3) agents often hold incorrect
  expectations about others' plans, observations, and communication.
  Besides these issues, through large-scale simulation, we  also observe rare but interesting emergent coordination behavior between agents including role division, resource division, and negotiation.
  Our research not only presents
  a novel benchmark for collaborative coding, but also calls for a research shift from pursuing individual
  agent capability to developing \emph{social intelligence}: the ability to understand others,
  communicate effectively, and coordinate actions. 
\end{abstract}

\begin{document}
  \maketitle

\subfile{sections/01-introduction}

\subfile{sections/02-cooperator_benchmark}
  \subfile{sections/03-experiment}
  \subfile{sections/04-cooperation}

  \subfile{sections/05-communication}

  \subfile{sections/06-coordination}

  \subfile{sections/07-related_work}

  \subfile{sections/07-conclusion}

  \section*{Acknowledgments}
This research is supported in part by grants from ONR grant N000142412532, NSF grant IIS-2247357, DSO National Laboratories (DSO), and support from SAP. We thank Google Cloud Platform and Modal Platform for their credits. We thank Yutong Zhang, Gavin Li, Hannah Cha, John Yang, Yijia Shao and all members of Stanford SALT Lab for their help and feedback throughout this project.

\input{main.bbl}
  \appendix
  \subfile{sections/08-appendix}

\end{document}

%% file: math_commands.tex
\usepackage{amsmath,amsfonts,bm}

\def\eqref#1{equation~\ref{#1}}

\def\1{\bm{1}}

\DeclareMathAlphabet{\mathsfit}{\encodingdefault}{\sfdefault}{m}{sl}
\SetMathAlphabet{\mathsfit}{bold}{\encodingdefault}{\sfdefault}{bx}{n}



%% file: sections/01-introduction.tex
\vspace{-30pt}

\section{Introduction}
\vspace{-5pt}
Most achievements in modern civilization arise from individuals working cooperatively, from the construction of cathedrals to the development of open-source software \citep{raymond1999cathedral,woolley2010evidence}.
In human societies, such cooperation relies on social intelligence: the ability to communicate intentions, understand others' goals, and negotiate mutually compatible solutions \citep{humphrey1976social}. 
This capability is often viewed as what makes us uniquely human and the basis of human thinking \citep{tomasello2014natural}.
As we deploy AI agents in cooperative settings, whether strong individual capabilities translate to effective cooperation with either humans or agents remains an open question.
In this paper, we empirically demonstrate that for current AI systems, there is a curse of coordination: \emph{agent cooperation perform much worse than a single agent given the same total workload.} 
This deficit presents a fundamental barrier to deploying AI systems that can work alongside humans or other agents.
We theorize that at a fundamental level, effective human–AI and agent–agent cooperation rely on the same coordination abilities. 

\begin{tcolorbox}[
  colback=gray!5,
  colframe=gray!50,
  arc=2pt,
  boxrule=0.5pt,
  left=6pt,
  right=6pt,
  top=4pt,
  bottom=4pt,
  title={\small\textbf{Glossary}},
  fonttitle=\sffamily
]
\small
\begin{description}[leftmargin=0pt, labelsep=0.5em, itemsep=2pt]
  \item[Cooperation:] When two or more agents/humans work together towards a shared goal, where an agent may altruistically help another achieve things outside their original responsibility.
  \item[Collaboration:] When two or more agents/humans work together towards a shared goal.
  \item[Coordination:] The capability to act and communicate in accordance with other agents/humans.
\end{description}
\end{tcolorbox}

Existing research on automating human tasks and multi-agent systems largely sidesteps this challenge by either providing more scaffolds \citep{fourney2024magenticonegeneralistmultiagentsolving, pan2025why, zhang2025agentorchestraorchestratinghierarchicalmultiagent, zhuge2024languageagentsoptimizablegraphs}, enforcing strict workflows \citep{hong2023metagpt, nguyen2024agilecoderdynamiccollaborativeagents, cheng2025hawkhierarchicalworkflowframework}, or providing active supervision and verification \citep{huang2025aiwork, xiang2025guardagentsafeguardllmagents, zheng2025webguardbuildinggeneralizableguardrail}. 
These systems rely on developer- or user-provided scaffolding to manage coordination, which limits flexible cooperation and places additional burden on humans.

We present \dataset, the first benchmark designed to measure how well agents can cooperate when handling individual tasks with potential conflicts. 
Considering software engineering as a realistic domain where humans typically need to navigate work in a team \citep{purna2011soft}, our benchmark offers verifiable evaluation for the success of agent cooperation. 
As illustrated in Fig. \ref{fig:teaser}, \dataset comprises 652 tasks constructed from 12 popular open-source libraries across Python, TypeScript, Go, and Rust. 
Eight co-authors of this paper with real-world software engineering backgrounds created new features, unit tests, and ground-truth code for these libraries, ensuring high-quality and realistic task design.

In \dataset, each task assigns each agent a feature to be implemented based on the same repository state. 
Conflicts are intentionally embedded at the code level, as the assigned features are logically compatible but require agents to modify overlapping or interdependent code.
For example, in Fig. \ref{fig:teaser}, one agent implements image mutability in the serialization process while another adds backup functionality to the same process. 
Without understanding each other's goals, plans, and expectations, their solutions may introduce incompatible changes. 
This mirrors real-world software development where coordination failures stem from insufficient mutual understanding. 
\dataset enables us to investigate three research questions:

\noindent\textbf{RQ1:} How well can agents cooperate with each other? (\S\ref{sec:cooperation_performance})

\noindent\textbf{RQ2:} What role does communication play in agent-agent cooperation? (\S\ref{sec:role_of_communication})

\noindent\textbf{RQ3:} What coordination failures do agents exhibit? (\S\ref{sec:coordination_failures})

Through evaluating state-of-the-art coding agents on \dataset, we observe \emph{the curse of coordination}: \texttt{GPT-5} and \texttt{Claude Sonnet 4.5} based agents achieve only 25\% with two-agent cooperation on CooperBench, which is around 50\% lower than a ``Solo'' baseline which uses one agent to implement both features. 

Diving deeper into the coordination failures, we identify three key issues. First, communication channels become jammed with vague, ill-timed, and inaccurate messages where agents fail to respond to direct questions, send messages that arrive too late to inform decisions, or flood channels with repetitive status updates that lack actionable detail.
Second, even with effective communication, agents deviate from their commitments. They make unverifiable claims about code state, ignore agreed-upon integration points, and break explicit promises.
Third, agents hold incorrect expectations about their partner's plans, observations and duplicate work despite warnings and overwrite changes they believe will merge cleanly (\S\ref{sec:coordination_failures}).

Besides failures, we are excited to report emergent coordination behaviors which often lead to the success of the \dataset tasks. These coordination behaviors are rarely performed by the agents, but through our large-scale simulation, we uncover three major categories of them: role division, resource division, and negotiations (\S\ref{sec:emergent-coordination-behavior}). These examples hint at a path of coordination capability acquisition through reinforcing success on \dataset. 

We contribute both a novel understanding of what agents need to become effective teammates and a practical benchmark for measuring progress. Our open-sourced \dataset platform enables researchers and practitioners to evaluate and improve cooperative coding agents.

%% file: sections/02-cooperator_benchmark.tex
\section{\dataset Benchmark} \label{sec:dataset}

\begin{wrapfigure}[23]{r}{0.5\linewidth}
    \centering
    \vspace{-45pt}
    \includegraphics[width=\linewidth]{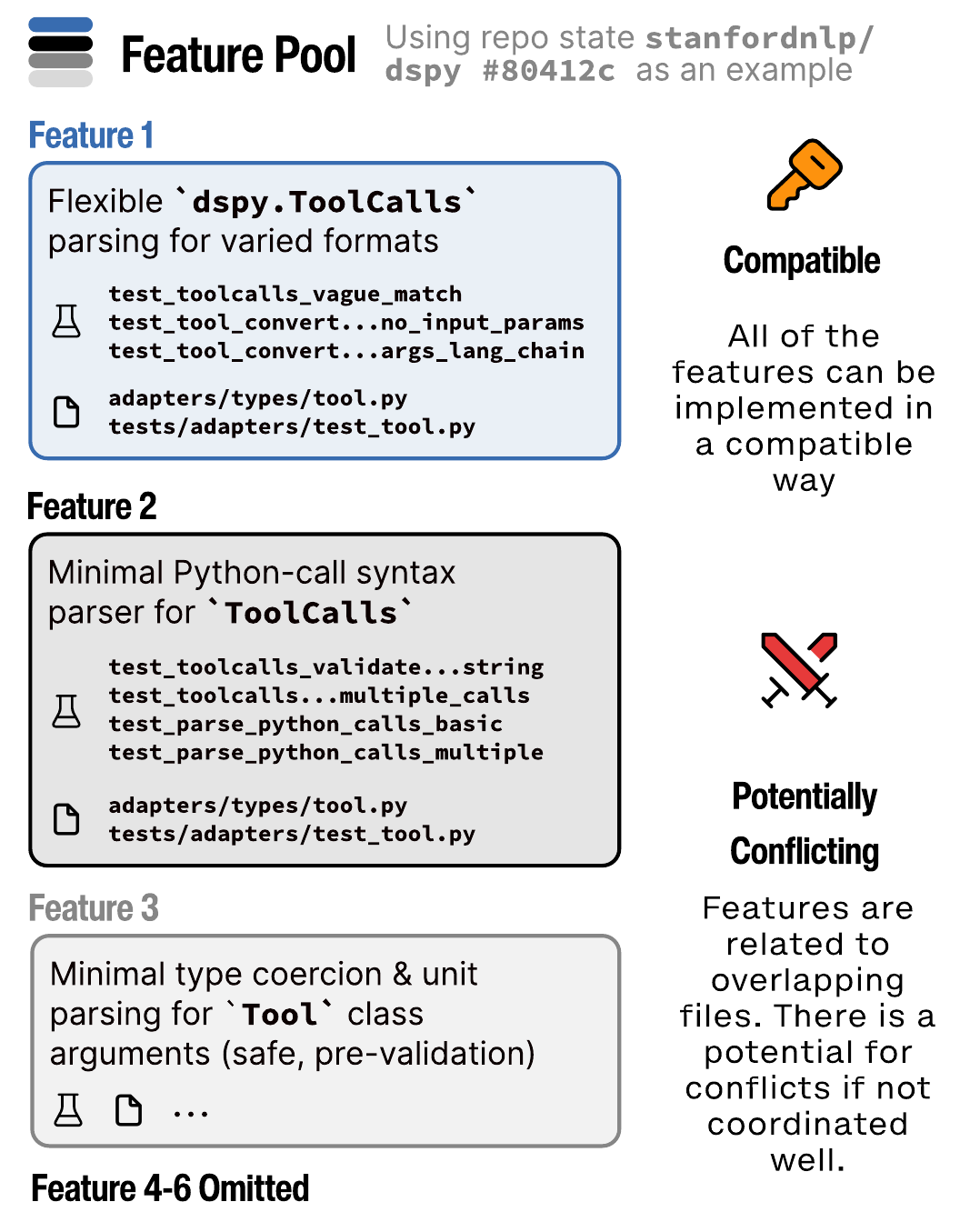}
    \caption{\small An example feature pool based on DSPy GitHub repository.
    This feature pool has 6 features which can be implemented compatibly based on the repository state, but without coordination agents could conflict with each other.}
    \label{fig:feature_pool}
\end{wrapfigure}

\dataset seeks to satisfy the following desiderata:
(1) \emph{Realism}: the tasks should be reasonable for a software development team to work on at a given repository state. 
(2) \emph{Conflict potential}: the agents' scopes should overlap with each other so that they need to coordinate well to avoid potential conflicts. 
(3) \emph{Verifiable}: the success of the tasks can be evaluated with a pipeline that is deterministic and interpretable. 
These three desiderata provide a basis for accurately measuring the real-world cooperation capabilities of agents. 

\subsection{Task space}\label{sec:task_space}

\noindent\textbf{Task} Each task consists of a repository state, two features, and two corresponding sets of unit tests. 
The two features are drawn from a pool of features (like the one illustrated in Fig. \ref{fig:feature_pool}) that can be simultaneously implemented on the given repository state. 
The patches from the two agents are merged and evaluated. 
Each agent's goal is to get their assigned feature implemented in the merged patch.\footnote{Agents have the freedom to redivide the two features as long as the merged patch implements both features.
Agents perform this kind of coordination occasionally well.
Check out \S\ref{sec:emergent-coordination-behavior} for concrete examples.}
Based on a pool of $n$ features, there will be $\binom{n}{2}$ tasks when we are evaluating agents self-play, and double the number when we evaluate two different agents cooperate with each other. 
Note that agents can only view their own features. 
For example, in Fig. \ref{fig:feature_pool}, there are 6 features in this pool, which produces 15 tasks for evaluating \texttt{GPT-5} agents cooperating with each other. 
If we want to evaluate how well \texttt{GPT-5} agents cooperate with \texttt{Claude Sonnet 4.5} agents, we will have 30 tasks drawn from this pool. 
In \dataset we have 34 such features pools.

\noindent\textbf{Features} In this paper, we use \texttt{features} to denote desirable changes to the codebase that implement missing functionality, fix existing bugs, or both. 
As illustrated in Fig. \ref{fig:feature_pool}, each \texttt{feature} is described in a markdown file, which includes a title, description, examples, and a list of files which may be relevant. 
For each feature, we write a list of unit tests without the help of coding assistants to ensure accurate evaluation of the implementation. 
In addition, we write a ground-truth solution to understand the potential conflicts between features and to verify that the given feature can be implemented on the repository and pass the unit tests. 
The tests and the ground-truth solution is not provided to the agents to prevent test leakage.

\noindent\textbf{Task composition} For each repository state, we create a pool of feature candidates. 
These features are \emph{compatible} and \emph{potentially conflicting}. 
``Compatible'' means the features can be implemented jointly. 
To verify this, we produce a joint ground-truth solution of all features in the pool, which passes all individual unit tests. 
``Potentially conflicting'' means the features have overlapping code logic changes that influence each other. 
In our dataset, 77.3\% of tasks have conflicting ground-truth solutions. 
As a result, \dataset tasks are not adversarial, but still require the capability to cooperate under conflicts by communicating individual goals, understanding others' plans, and negotiating mutually compatible solutions. 

\noindent\textbf{Action space} Agents can take two kinds of actions in real time: the \emph{communication tool} and \emph{computer-use tools}. 
The communication tool allows agents to send open-ended natural language messages to each other, and the computer-use tools include file and terminal operations. 
In our paper, we limit the computer-use tools to local operations to control the experiments. 
In the future, researchers could consider GUI and browser-based actions to expand the tasks the agents can take. 
Both agents can use these tools at any time, without synchronizing their turns with each other. 
This not only raises the flexibility of agents, but also poses challenges for agents to timely communicate and execute commands. 
In our benchmarking process, we use cloud virtual machines for agents to ensure isolated workspaces and sufficient resources. 
We set an upper-bound number (100)\footnote{We do not observe performance gains on our tasks from raising this number.} of actions an agent can take to complete the tasks. 
  
\subsection{Evaluation pipeline}\label{sec:evaluation_pipeline}

Cooperation is hard to evaluate, but we make the product of the cooperation verifiable.
\dataset evaluates tasks based on two criteria: (1) \emph{compatible solutions} and (2) \emph{implementation correctness}.

\noindent\textbf{Solution compatibility} After the two agents complete execution, we attempt to merge their resulting patches using \texttt{git merge-file -L patch\_1 -L repo\_state -L patch\_2}.
This operation captures whether the independently produced solutions are structurally compatible. 
In practice, some merge failures arise from superficial differences such as formatting or indentation styles (e.g., K\&R versus Allman) rather than substantive conflicts. 
To avoid treating such cases as coordination failures, we train a small coding model (\texttt{Qwen 3 Coder 1.5B}; \citealt{yang2025qwen3technicalreport}) on synthetic examples to resolve trivial merge conflicts when standard merging fails.
This step ensures that the compatibility check reflects semantic agreement between solutions rather than low-level stylistic discrepancies, while leaving the overall cooperation score largely unaffected (App. \S~\ref{app:resolver}).
If even the coding model cannot produce a patch without conflicts, the agents both fail the tasks.

\noindent\textbf{Implementation correctness} If we successfully merge the two patches into the repository state, we run both sets of unit tests on the merged codebase. 
As mentioned before, we do not restrict agents to only finish their own work. 
If they can coordinate well, they can divide their two features in a different way as long as the merged solution can pass the two features' tests. 
This evaluation pipeline ensures a rigorous evaluation of the cooperation outcome.

\subsection{Dataset Construction}
\label{sec:dataset_construction}

\begin{figure}[htbp]
    \centering
    \includegraphics[width=\linewidth]{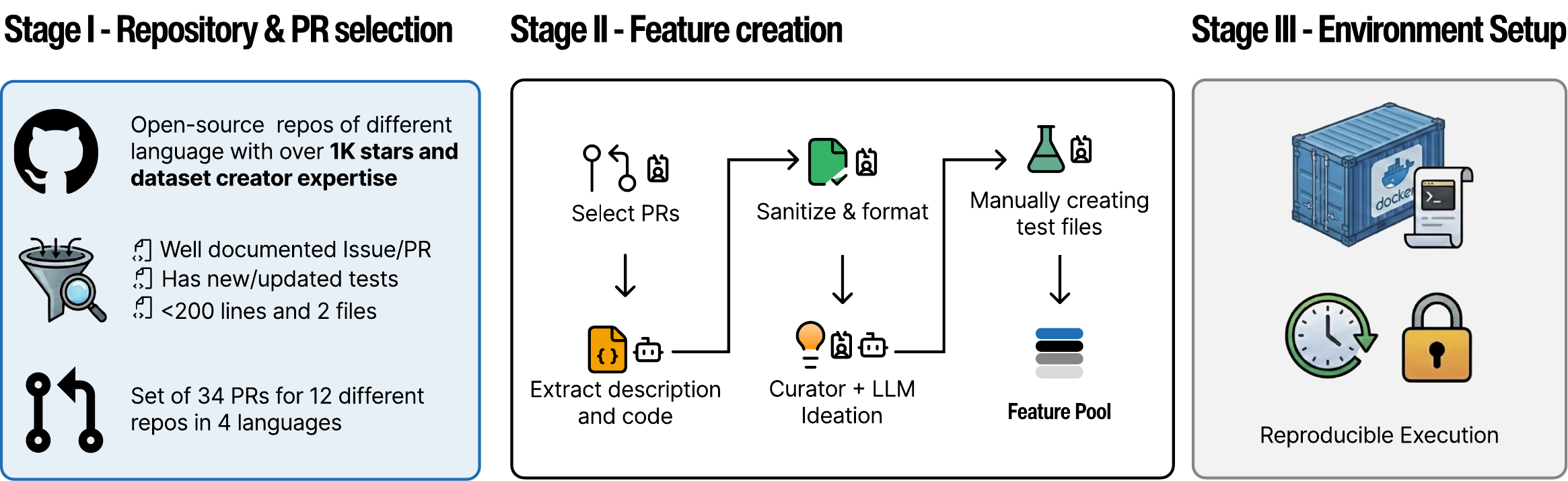}
    \caption{The \dataset construction pipeline.
    Each task is
    carefully engineered by domain experts to ensure conflicts are realistic, resolvable, and
    representative of production software development challenges.}
    \label{fig:dataset_construction}
\end{figure}

\dataset is constructed through a three-stage pipeline that grounds tasks in real software development and enables controlled evaluation of coordination (Fig. \ref{fig:dataset_construction}). 
To create the pools of features, we start from real-world feature implementations and proceed as follows: 
(Stage I) we write \textit{anchor features} drawn from popular repositories, each of them is a slight modification of a real pull request (PR) authored by human contributors; 
(Stage II) for each anchor feature, we expand the pool by introducing a family of \textit{adjacent features} authored by human annotators, representing plausible alternative features that could realistically co-occur; and 
(Stage III) we validate the compatibility of each feature pool by executing and testing all feature combinations in a controlled environment to rule out intrinsically incompatible specifications.

\noindent \textbf{Stage I: Repository and PR Selection} 
In the first stage we select twelve actively maintained open-source repositories spanning Python, TypeScript, Rust, and Go. 
Each repository exceeds one thousand GitHub stars and does not appear in SWE-Bench~\citep{jimenez2023swe} or Multi-SWE-Bench~\citep{zan2025multiswebench}, reducing data contamination risk. 
Selection is guided by curator expertise so that each repository is assigned to an author familiar with its architecture and development practices. 
We extract PRs that meet strict inclusion constraints: clear feature description, code+tests, feature addition, bounded change size, and robust tests. 
Appendix~\ref{app:dataset_stats} provides full selection details and thresholds, and App. Tab.~\ref{tab:repos} summarizes the repository distribution.

\noindent\textbf{Stage II: Feature Extraction and Augmentation}
In the second stage, we convert each selected PR into a feature pool containing one anchor feature and multiple synthetic adjacent features. 
We sanitize and rewrite original PR descriptions into self-contained specifications to prevent information leakage. 
Curators author adjacent features to plausibly co-occur and to create natural overlap without adversarial specifications (with LLM-assisted ideation). 
Appendix~\ref{app:dataset_stats} provides full details on adjacent-feature design, manual test writing, and gold-solution validation. All features derived from the same base commit constitute a feature pool with two to 12 features.
To ensure the compatibility among all features in a pool, we construct a single gold patch that jointly implements all features in each set and passes all associated tests.

\paragraph{Stage III: Environment and Reproducibility}

The final stage provides a deterministic execution environment for evaluating agents. 
Each task set includes an automated setup script that clones the repository at the exact base commit, installs dependencies, and executes the full test suite to verify the environment. 
To ensure consistent behavior across hardware and operating systems, we additionally provide containerized environments that encapsulate the complete repository state and all runtime dependencies. 
These environments guarantee reproducible execution and isolate agent behavior from external variability, enabling reliable measurement of coordination performance through the evaluation pipeline described in \S\ref{sec:evaluation_pipeline}. 

Dataset composition and feature-complexity statistics are reported in App.~\ref{app:dataset_stats}. 
Together, these findings demonstrate that \dataset features are individually tractable and realistic, ensuring that the benchmark's primary challenge arises from coordinating partially overlapping implementations rather than from executing unusually complex or oversized programming tasks.

%% file: sections/03-experiment.tex
\section{Experiment Settings}\label{sec:experimental_design}


\dataset allows us to study the following research questions. First, how well can current state-of-the-art foundation models cooperate with each other when they are used in coding agents? Second, do agents use the communication channel effectively for coordination? And, what are the reasons why agents fail or succeed on \dataset? 

In order to evaluate models fairly, we create an agent framework incorporating leading open-source coding agent framework \texttt{OpenHands (v0.54)}~\citep{wang2024openhands}. The two agents perform their own work in their respective docker-based containers without interruption from another agent. Since OpenHands was not designed as a framework which performs multi-agent cooperation, we created a communication tool (\S\ref{sec:task_space}) using an SQL database for message passing. This communication tool supports message sending action. When an agent sends a message to another agent, the other agent will immediately receive it, and include it in the prompt of the next step. This communication setting achieves both real-time communication and asynchronous execution. We open-source this framework to not only ensure reproducibility of our experiments, but also provide a starting point for researchers to build multi-agent cooperation systems which can perform multiple tasks and resolve conflicts. 

\emph{However, note that \dataset does not tie with the agent framework or the communication tool.} In this paper, we are more concerned with foundation models' intrinsic capability to cooperate, so we do not compare different agent frameworks or creative methods to enhance coordination. In the future, researchers should use \dataset to compare different models, different frameworks, and different combinations as well. We especially encourage researchers to develop novel frameworks or to train agents to achieve higher Coop scores or to close the Solo-Coop gaps (\S\ref{sec:cooperation_performance}) on \dataset.
Similarly, we encourage researchers to develop other communication tools, e.g. screen sharing, to expand the communication bandwidth or reduce the communication noises. 

We evaluate the performance of five language models, both closed-source ones, and open-source ones: \texttt{GPT-5}, \texttt{Claude 4.5 Sonnet}, \texttt{MiniMax-M2}, \texttt{Qwen3-Coder-30B-A3B-Instruct}, and \\\texttt{Qwen3-30B-A3B-Instruct-2507}.
We serve the two Qwen models via vLLM\footnote{\url{https://vllm.ai/}}, GPT-5 and Minimax models via their respective official API, and the Claude model through GCP. 

%% file: sections/04-cooperation.tex
\section{How well are agents able to cooperate with each other?}\label{sec:cooperation_performance}

In \dataset, each of the two agents are assigned a feature to implement, which will be called the Coop setting to distinguish from the Solo baseline. In the Solo baseline, the two tasks are assigned to one agent. For humans, teams should perform better or faster than individuals, which is the bottom line for cooperation to be considered as functional. We hypothesize for agents, the advantage of cooperation is overwhelmed by their incapability to coordination. This should lead to a ``coordination gap'': two agents perform worse than one agent for the same workload. 

  \begin{figure}[!t]
    \centering
    \includegraphics[width=\linewidth]{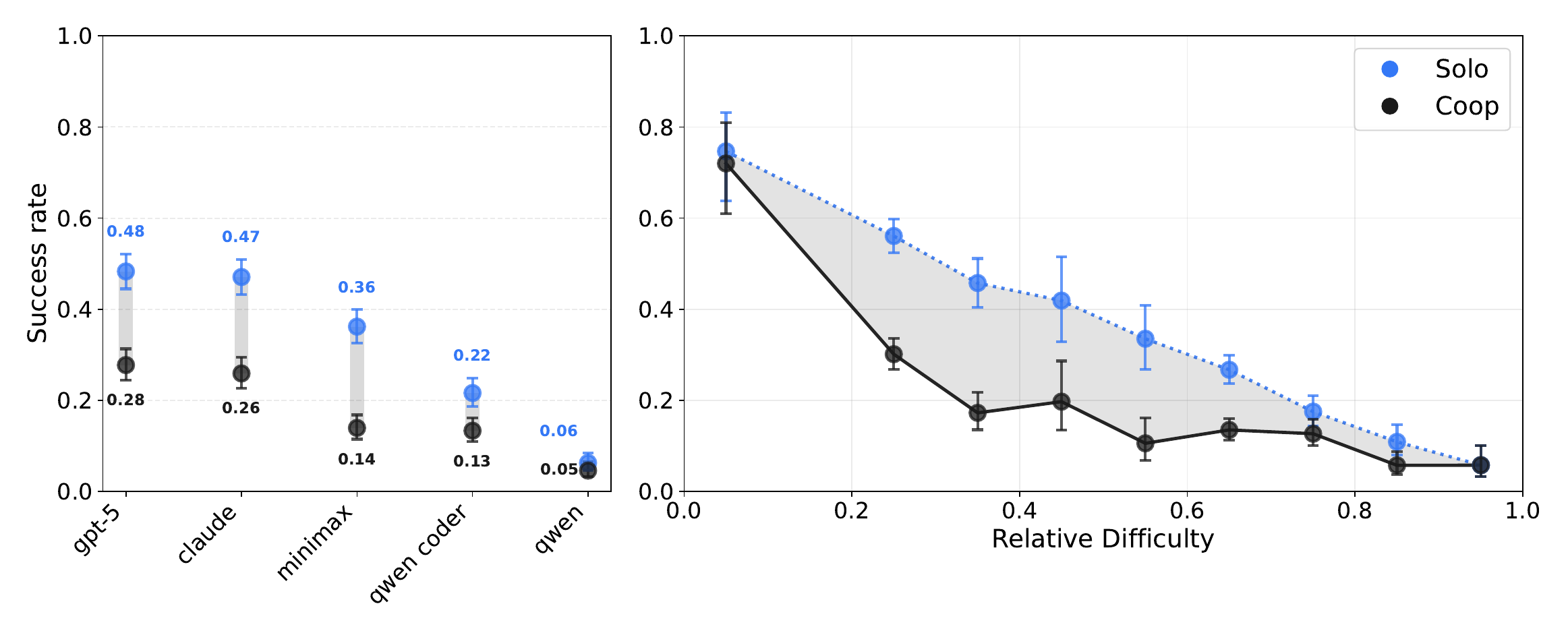}
    \caption{
    \textbf{Left:} Under Coop setting, agents with different foundation models perform significantly worse than how they perform under Solo setting, except for \texttt{Qwen3-30B-A3B-Instruct-2507}, which performs bad under both settings. This Solo-Coop gap is what we call the ``coordination gap''. \textbf{Right:} The relationship between tasks' \emph{technical} difficulties and Solo-Coop gap. The shaded area has a large middle section which shows that the coordination gap is larger for middle-level tasks than for tasks which are extremely easy or difficult.}\label{fig:main_results}
  \end{figure}

\paragraph{The curse of coordination.} As shown in Fig.~\ref{fig:main_results} (Left), 
across all models, success rates under the Coop setting is consistently lower than those under Solo settings, which means when two agents need to coordinate between them, they perform even worse than one agent ``solo''ing the two features. 
This coordination gap is as large as 50\% in the leading models: \texttt{GPT-5}, \texttt{Claude Sonnet 4.5}, and \texttt{Minimax M2}. Qwen models have smaller gaps, but their Solo setting score is much lower as well. All error bars in Fig.~\ref{fig:main_results} are 95\% \emph{Wilson} confidence intervals computed over task sets (App.~\ref{app:ci}).

\paragraph{Mid-difficulty crisis.}
As shown in Fig.~\ref{fig:main_results} (Right), the gap between the two settings is larger and more significant on the tasks with middle-level technical difficulty than on the ones which are too easy or too hard. 
Here we stratify tasks by \textit{relative difficulty}.
For each task pair \(t\), we define a raw difficulty score
\(d(t) = 1 - \frac{1}{|\mathcal{M}|}\sum_{m \in \mathcal{M}} \mathrm{Solo}_m(t)\), where \(\mathrm{Solo}_m(t)\) denotes model \(m\)'s Solo success on \(t\).
For visualization, we linearly rescale \(d(t)\) to \(\tilde d(t) \in [0,1]\) using the minimum and maximum \(d(t)\) values in the benchmark.
We bucket tasks by \(\tilde d(t)\) and report success rates as a function of \(\tilde d(t)\) for both Solo and Coop.
This result points out that agents struggle to balance the two pressures for technical difficulty and cooperation difficulty. 
When tasks are too easy, agents could spare more effort for coordination, but when tasks are getting harder, agents cannot effectively coordinate.

\paragraph{Scaling the number of cooperating agents.}
Our hypothesis is that increasing the number of agents in the same cooperative workspace exacerbates coordination overhead (e.g., more context to track and more opportunities for inconsistent plans), leading to lower end-to-end success.
To probe this directly, we run a small scale experiment using 46 tasks from 3 separate task sets where we scale the number of concurrently cooperating agents from 2 to 4 while keeping the cooperative setting fixed.
We observe a monotonic decline in success as the number of agents increases. Specifically, performance drops from \(68.6\%\) with 2 agents to \(46.5\%\) with 3 agents and further to \(30.0\%\) with 4 agents on the tasks, reinforcing the ``curse of coordination'' beyond the 2-agent setting.


%% file: sections/05-communication.tex
\section{What is the role of communication in agent-agent cooperation?}\label{sec:role_of_communication}

In \dataset, the communication tool we provide is the only channel agents could use to coordinate with each other. Can agents effectively use it? We hypothesize that although agents might actively use the tool, their communication might be far from effective or efficient. To evaluate this, we compare with a baseline setting, where the communication tool is banned, i.e. ``no comm''.

\begin{figure}[!t]
    \centering
      \centering
      \includegraphics[width=\linewidth]{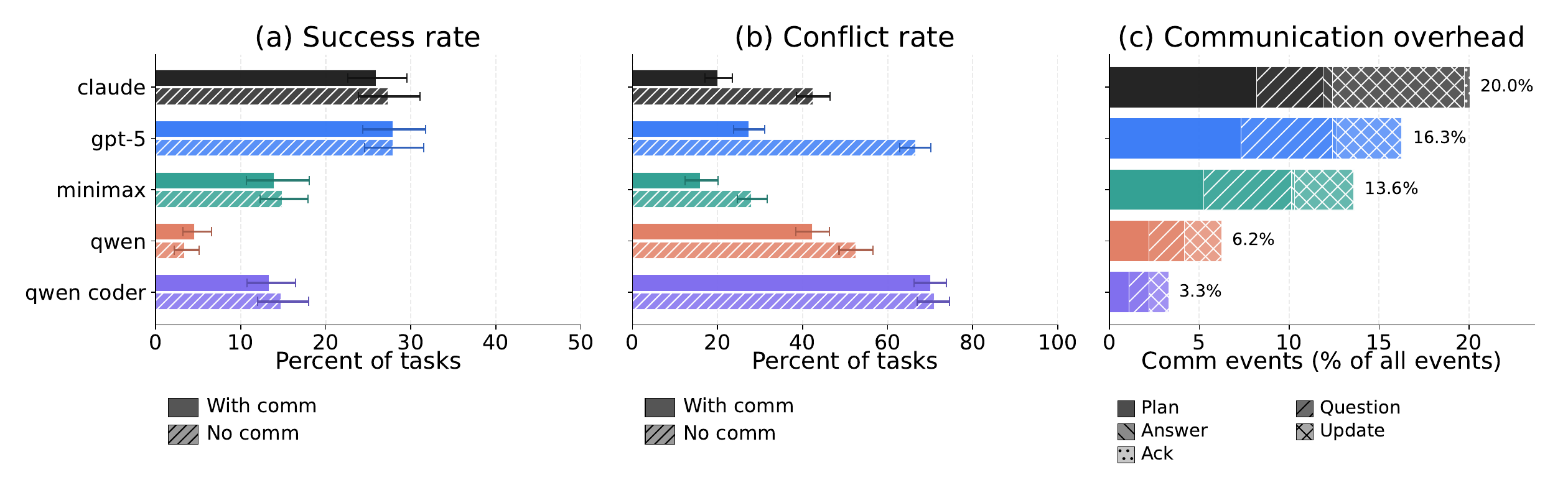}
    \caption{\textbf{(a)} Effect of inter-agent communication on cooperation success or lack thereof. All agents fail to use communication for improving cooperation success. \textbf{(b)} Communication substantially reduces naive merge conflicts across all models. \textbf{(c)} Communication overhead as a percentage of all execution events, broken down by message type. Models that communicate more (e.g., \texttt{Claude Sonnet 4.5}, \texttt{GPT-5}) show larger reductions in conflict rate.}\label{fig_comm_conflict_and_overhead}

\end{figure}
\paragraph{Communication does not lead to better cooperation.}
As shown in Fig.~\ref{fig_comm_conflict_and_overhead} (a), none of the models effectively leverage communication tool to achieve higher cooperation success. The difference between ``with comm'' and ``no comm'' settings is not statistically significant. This shows that existence of the communication tool does not help coordination. Does this mean agents are not using them? We quickly negate this question through examine the usage and the conflict rate. 

\paragraph{Communication reduces merge conflicts.}
As shown in Fig.~\ref{fig_comm_conflict_and_overhead} (b), communication does significantly reduce the merge conflicts between patches in \texttt{Claude Sonnet 4.5}, \texttt{GPT-5}, \texttt{MiniMax M2}, and \texttt{Qwen Instruct}. This shows that agents could leverage the communication tool to reduce overlap in their work, despite that just avoiding conflicts does not warrant cooperation success. 
Communication also consumes a meaningful share of the agent's action budget.
Fig.~\ref{fig_comm_conflict_and_overhead} (c) reports the frequency of all communication speech act types.
This result shows that agents spent as much as 20\% of the steps in communication, within which planning, questioning, and updating each almost takes up 1/3 of steps. But why this much effort in communication does not translate into better cooperation?

\paragraph{What distinguishes effective communication?}
To understand why communication helps conflicts but not success, we analyze what \emph{successful} communication looks like. 
Three patterns emerge.


\noindent\textit{First, successful agents plan more and question less.}
Trajectories that avoid conflicts have a Plan:Question ratio of 2.04, compared to 1.31 for conflict trajectories. This suggests that questions are a \emph{symptom} of coordination problems, not a cure. Agents that are already struggling tend to ask more questions, but questioning does not prevent conflicts.

\noindent\textit{Second, first-turn planning is the strongest predictor.}
Having a \texttt{Plan} message in the very first turn nearly halves the conflict rate (29.4\% vs 51.5\%). This effect is robust across difficulty levels: in 7 out of 8 difficulty buckets, first-turn planning significantly reduces conflicts, with the effect actually \emph{stronger} for harder tasks (39\% reduction at the highest difficulty).

\noindent\textit{Third, specificity matters.}
Successful trajectories contain significantly more concrete references: 32.6 line number mentions versus 22.5, and 13.1 file path mentions versus 10.0. Agents that communicate \emph{where} they are editing with specific line ranges successfully avoid overlapping changes.

\paragraph{Spatial vs.\ semantic coordination.}
These findings explain why communication helps conflicts but not success. Merge conflicts are fundamentally a \emph{spatial} coordination problem: agents must agree on who edits which lines. The patterns above (early planning, specific line numbers, file paths) all address spatial coordination, and they work.

However, task success requires \emph{semantic} coordination: understanding \emph{what} to implement, not just \emph{where}. Our case study in Appendix~\ref{app:case_study} illustrates this gap. Two agents successfully coordinated on line numbers and edit ranges (spatial), yet failed because they never discussed the actual parameter values their implementations should use (semantic). They solved the ``formatting'' problem of avoiding overlapping edits but not the ``design'' problem of ensuring compatible implementations.

\begin{figure}[!t]
    \centering
    \includegraphics[width=\linewidth]{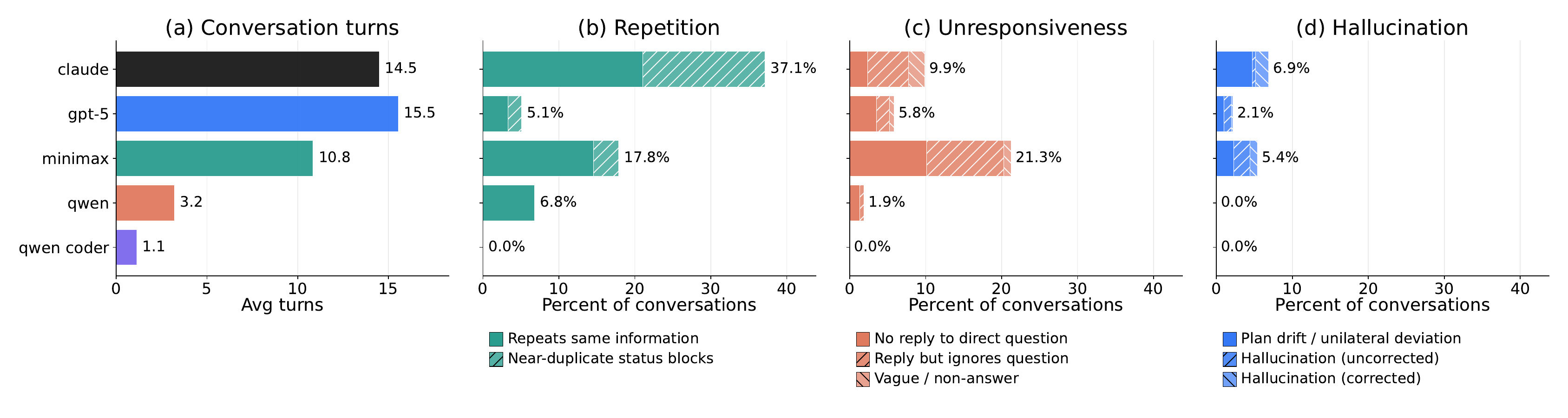}
    \caption{Break down frequencies of different kinds of communication errors.}\label{fig_comm_errors}
\end{figure}

\paragraph{Repetition, Unresponsiveness, and Hallucination.} Beyond the spatial-semantic gap, the communication itself is often flawed.
We identify three major communication problems, and show their frequencies in Fig.~\ref{fig_comm_errors}.
We automatically detect these patterns using an LLM-as-judge approach with a precision-focused taxonomy; see Appendix~\ref{app:comm_error_detection} for the full rubric and evidence requirements.
Repetition consumes budget without adding constraints a partner can act on, which is consistent with high communication overhead without commensurate gains in end-to-end success.
Unresponsiveness breaks the feedback loop when one agent asks for a decision that gates implementation, and incorrectness creates false shared context, such as asserting an interface decision or a completed change that is not actually satisfied.
Hallucination results in noises which makes it hard to partners to coordinate under imperfect information.

In this section, we show that the communication tool is heavily used, but \emph{not} properly leveraged by agents for coordination.
This shows that agents lack the critical \emph{pragmatics} understanding of language: communication is not just about message passing, but about achieving certain functions through passing the messages.
Agents are ``talking'' a lot, but they cannot achieve their communication goals through communication when communication channel is jammed with repetitions, unresponded questions, or false information.

%% file: sections/06-coordination.tex
\section{What are the coordination failures that the agents exhibit?}\label{sec:coordination_failures}

\definecolor{info}{RGB}{220,235,255}
\definecolor{integ}{RGB}{255,235,220}
\definecolor{struct}{RGB}{220,255,220}
\definecolor{comm}{RGB}{245,220,245}

\newcommand{\emojiicon}[1]{%
  \raisebox{-0.05em}{\includegraphics[height=1.15em]{#1}}%
}

\newcommand{\rootcell}[3]{%
  \parbox[t]{0.23\textwidth}{%
    \raggedright%
    \hspace{0.6pt}\strut%
    \textbf{\emojiicon{#1}\,#2}\par%
    \if\relax\detokenize{#3}\relax\else
      {\scriptsize\textit{#3}}\par%
    \fi
    \strut%
  }%
}

Section~\ref{sec:role_of_communication} showed that communication alone does not improve coordination. Why not? We find that even when agents communicate their plans, they struggle to honor commitments and anticipate partner actions. Coordination failures stem from three capability gaps: \emph{communication} (failing to exchange key information), \emph{commitment} (not following through on promises), and \emph{expectation} (failing to model what partners are doing). We first categorize failures by their observable \emph{symptoms} (\S\ref{sec:failure_syptoms}), then identify these underlying \emph{causes} (\S\ref{sec:failure_reasons}).

\input{sections/06-symptoms_table}

\subsection{Failure Symptoms}
\label{sec:failure_syptoms}
We analyze all failed Coop trajectories across all five models on the full dataset. Through iterative qualitative coding, we develop the failure symptom taxonomy shown in Tab.~\ref{tab:failure_symptoms}. We then use \texttt{GPT-5} as an LLM-as-a-Judge to categorize trajectories at scale, yielding the frequency distribution in Tab.~\ref{tab:failure_symptoms}. The resulting vocabulary provides a structured way to diagnose coordination breakdowns. See App.~\ref{app:symptom_annotation} for the annotation procedure and human validation.

\subsection{Failure Reasons}
\label{sec:failure_reasons}

\input{sections/06-causes_table}

Symptoms describe \emph{what} went wrong; causes explain \emph{why}. To identify the underlying capability gaps, we manually reviewed 50 failed Coop traces. For each trace, we examined the symptom labels, conversation logs, and merged artifacts to determine why coordination broke down. We grouped root causes into three categories shown in Tab.~\ref{tab:failure_causes}. Unlike symptoms, which can be reliably detected by an LLM annotator, causes require deeper interpretation of the coordination dynamics and are therefore manually assigned.

\input{sections/06-coordination-cards}

\medskip
The examples above reveal why coordination, rather than raw coding ability, is often the limiting factor. The common thread is \emph{partial observability}. Each agent acts while holding an uncertain model of its partner's state, edits, and commitments. A merge can be conflict-free yet still embed incompatible assumptions.

These causes manifest through the symptoms in Tab.~\ref{tab:failure_symptoms}. Expectation failures produce work overlap and silent overwrites, commitment failures lead to unverifiable claims and broken promises, and communication failures result in unresponsiveness and repetition.

These failures suggest current models lack reliable representations for (i) \emph{partner state} (what the other agent has actually changed), (ii) \emph{checkable commitments} (contracts verifiable after merge), and (iii) \emph{cross-branch integration reasoning} (anticipating how independent patches interact). Coordination requires more than plausible code. It requires \emph{verifiable} and \emph{actionable} constraints for a partner operating under isolation. This explains why prompt optimization yields only marginal improvements (App.~\ref{app:prompt_optimization}). Most errors stem from coordination challenges, not prompt wording.

\noindent\textbf{The trust paradox.} We hypothesize that a deeper tension underlies expectation failures. Models are trained to be cautious, requiring observable evidence and resisting unverifiable assertions. This is a sensible default for single-agent interactions, where users may attempt to mislead the model. However, collaboration under workspace isolation requires the opposite. Agents must trust partner claims about states they cannot observe. When Agent~A reports ``I added the handler at line~50,'' Agent~B's instinct is to verify, but verification fails because they are on separate branches. This mismatch between \emph{verification-first} training and \emph{trust-requiring} collaboration may partly explain why agents consistently fail to update their model of partner state despite explicit communication.

Effective collaboration likely requires lightweight mechanisms that turn conversation into verifiable shared state, such as pasted signatures, explicit insertion-point contracts, and integration checks before declaring safety. We now turn to successful cases to see what these mechanisms look like in practice.

\subsection{Emergent Coordination Behavior}
\label{sec:emergent-coordination-behavior}
Among successful runs, we observe coordination patterns that are largely absent from failures. These behaviors are not prompted or scaffolded; they emerge when agents successfully navigate partial observability. What they share is a shift from vague intentions to specific commitments that a partner can verify even without seeing the underlying work. We identify three such patterns.
\vspace{-1em}
\paragraph{Role division.}
Agents agree on who handles which part of the task and establish clear boundaries around their scope.

\begin{figure}[htbp]
    \centering
    \includegraphics[width=\linewidth]{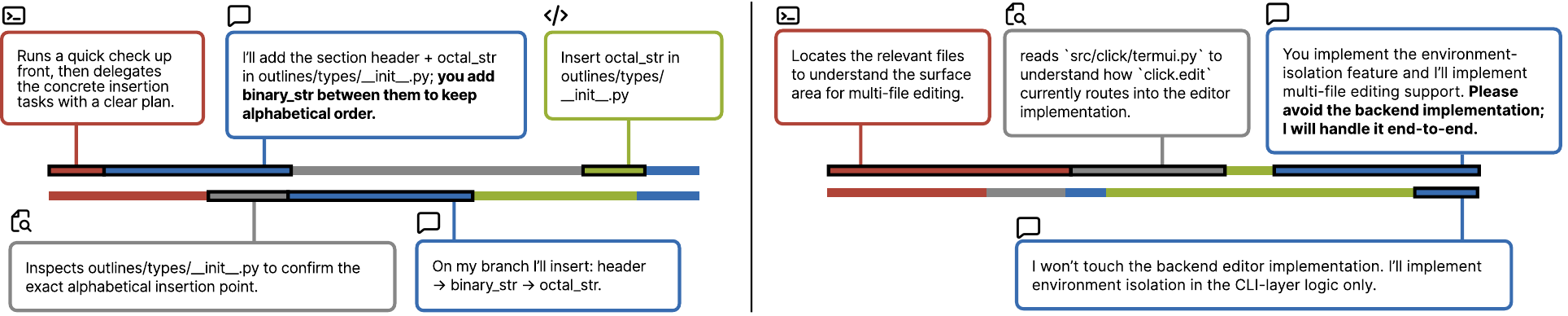}
\end{figure}

What distinguishes successful role division is mutual confirmation.
Under partial observability, a unilateral declaration can easily be missed or misunderstood.
When both agents explicitly acknowledge the split, they create verified shared understanding that both sides can rely on during independent work.
\vspace{-1em}
\paragraph{Resource division.}
Agents avoid collisions by partitioning shared resources, most commonly specific files, code ranges, or \textit{ownership blocks}.

\begin{figure}[htbp]
    \centering
    \includegraphics[width=\linewidth]{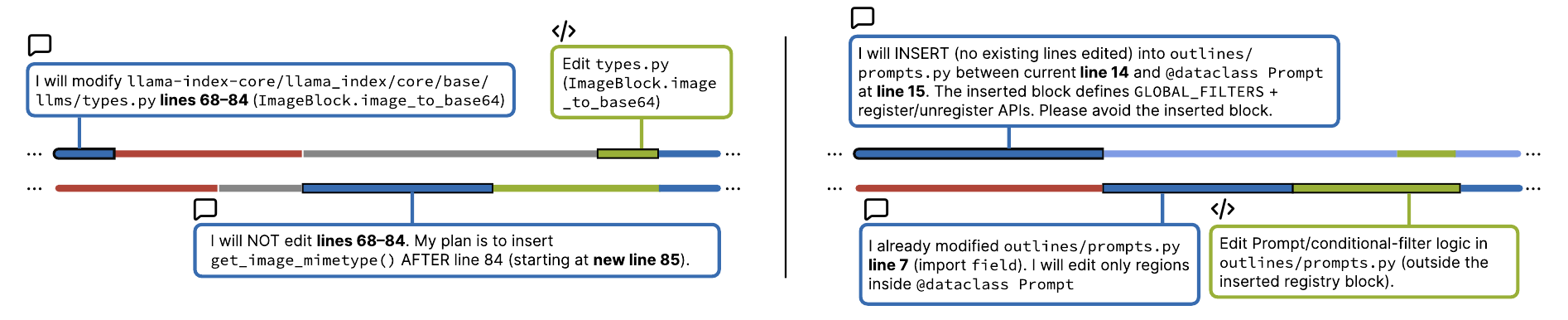}
\end{figure}

What makes resource division effective is specificity.
Vague commitments cannot be verified and thus require trust.
Line-level boundaries, by contrast, create safe zones where conflict is impossible by construction.
\vspace{-1em}
\paragraph{Negotiation.}
Agents resolve conflicting approaches by proposing alternatives and converging on a single plan before acting.

\begin{figure}[htbp]
    \centering
    \includegraphics[width=\linewidth]{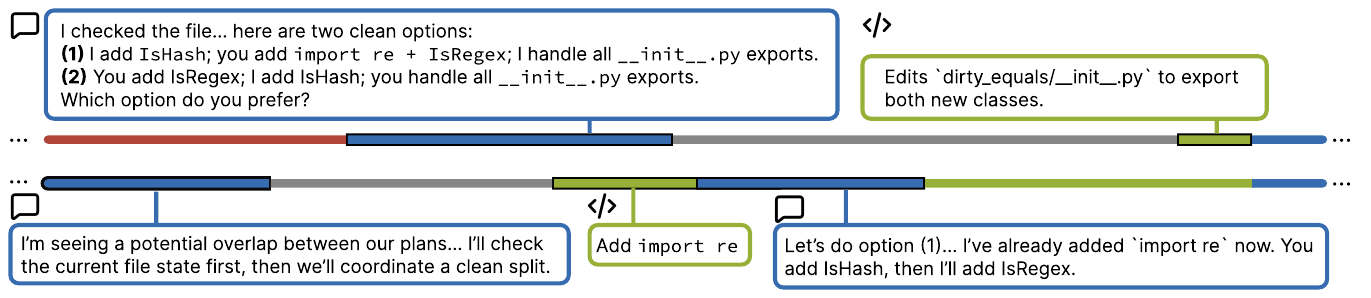}
\end{figure}

Effective negotiation does cognitive work for both parties.
By proposing mutually exclusive options that fully specify what each agent will do, one agent reduces a complex coordination problem to a simple choice.
The result is not just agreement on intent but complete action specifications that leave nothing to interpret.

These coordination patterns are rare in our traces but their presence in successful cases suggests that the underlying capability exists.
The challenge is not teaching agents new coordination skills but making existing ones reliable.

%% file: sections/06-symptoms_table.tex
\begin{table*}[t]
  \centering
  \small
  \caption{Coordination failure symptoms. Observable patterns in how coordination breakdowns surface in merged artifacts.}
  \label{tab:failure_symptoms}

  \begingroup
  \setlength{\tabcolsep}{4pt}
  \setlength{\extrarowheight}{1.5pt}
  \renewcommand{\arraystretch}{1.25}

\begin{tabular}{
  p{0.20\textwidth}
  p{0.70\textwidth}
  c
}
    \toprule
    \textbf{Symptom} & \textbf{Meaning} & \textbf{\%} \\
    \midrule
    Work overlap & Both agents independently implement the same functionality, duplicating work and overwriting details. & 33.2 \\
    Divergent architecture & Incompatible design decisions lead to semantic loss even under a clean merge. & 29.7 \\
    Repetition & Verbose status messages add little new information and reduce signal. & 14.7 \\
    Unresponsiveness & Direct questions or requests are not answered, breaking the decision loop. & 8.7 \\
    Unverifiable claims & Agent asserts a change or interface decision without evidence the partner can check (no checkable commitment). & 4.3 \\
    Broken commitment & Confident completion claims create false shared context when the promised change is absent. & 3.7 \\
    Dependency access & Missing risk communication leaves agents unable to anticipate merged dependency interactions (e.g., circular imports). & 1.7 \\
    Placeholder misuse & An explicit integration contract exists but is applied differently than agreed. & 1.5 \\
    Parameter flow & Ambiguity about a changing interface leaves one agent implementing against an outdated contract. & 1.3 \\
    Timing dependency & Agents agree on order but fail to communicate an enforceable plan that preserves it after merge. & 1.1 \\
    \bottomrule
  \end{tabular}
  \endgroup
\end{table*}

%% file: sections/06-causes_table.tex
\begin{table*}[t]
  \centering
  \small
  \caption{Coordination capability gaps. Underlying causes inferred through qualitative analysis of failure traces.}
  \label{tab:failure_causes}

  \begingroup
  \setlength{\tabcolsep}{4pt}
  \setlength{\extrarowheight}{1.5pt}
  \renewcommand{\arraystretch}{1.25}

  \begin{tabular}{
    @{\hspace{2pt}}
    >{\raggedright\arraybackslash}p{0.2\textwidth}
    >{\raggedright\arraybackslash\footnotesize}p{0.70\textwidth}
    >{\centering\arraybackslash}p{0.05\textwidth}
    @{}
  }
    \toprule
    \textbf{Cause} & \textbf{Definition} & \textbf{\%} \\
    \midrule
    \rowcolor{info}
    \rootcell{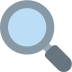}{Expectation}{} & Cases where one agent has clearly communicated what they are doing, but the other agent still treats the situation as if that work is not being done. This reflects a failure to model the state of the other agent's code changes and what that means for the system as a whole. & 42 \\
    \midrule
    \rowcolor{integ}
    \rootcell{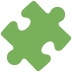}{Commitment}{} & Cases where an agent is not doing the things they promised to do. This includes failures to establish or maintain verifiable integration contracts, where agents make commitments but do not follow through on them. & 32 \\
    \midrule
    \rowcolor{comm}
    \rootcell{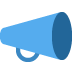}{Communication}{} & Breakdowns in using language to coordinate. This includes failures in information sharing and decision loops between agents, where agents do not effectively communicate their intentions, questions, or status updates. & 26 \\
    \bottomrule
  \end{tabular}
  \endgroup
\end{table*}

%% file: sections/06-coordination-cards.tex
\tcbset{
  failurecard/.style={
    enhanced,
    arc=2mm,
    boxrule=0.45pt,
    colframe=black!10,
    left=6pt,
    right=6pt,
    top=6pt,
    bottom=6pt,
  }
}

\tcbset{
  failuregroup/.style={
    failurecard,
    coltitle=black,
    fonttitle=\bfseries\color{black},
    title filled=false,
    valign=top,
    breakable,
    width=\textwidth,
  },
  subfailure/.style={
    enhanced,
    arc=1.6mm,
    boxrule=0.35pt,
    colframe=black!12,
    colback=white,
    left=6pt,
    right=6pt,
    top=6pt,
    bottom=6pt,
  }
}

\newcommand{\causefigure}[1]{%
  \begin{tcolorbox}[subfailure]
    \centering
    \includegraphics[width=\linewidth]{#1}
  \end{tcolorbox}%
}

\subsection{Representative examples of capability gaps}
We provide one representative example for each coordination capability gap. Additional symptom-level examples are available in Appendix~\ref{app:symptom_examples}.

\paragraph{Expectation.} In the first example, Agent A announces it will modify \texttt{prompts.py} and call B's \texttt{get\_global\_filters()}. Agent B states it will insert \texttt{GLOBAL\_FILTERS} at a specific location. Both agents communicate their plans explicitly, yet the merge fails. The problem is not missing information but failure to \emph{integrate} it. Despite hearing B's plan, A proceeds as if B's code won't exist. This is the most common cause, reflecting a fundamental difficulty in maintaining an accurate model of partner state during independent work.

\vspace{0.35em}
\begin{tcolorbox}[failuregroup,colback=info,
  title={\emojiicon{figs/emoji/1f50d.png}\ \textbf{Expectation}: Failure to model partner state}]
  \causefigure{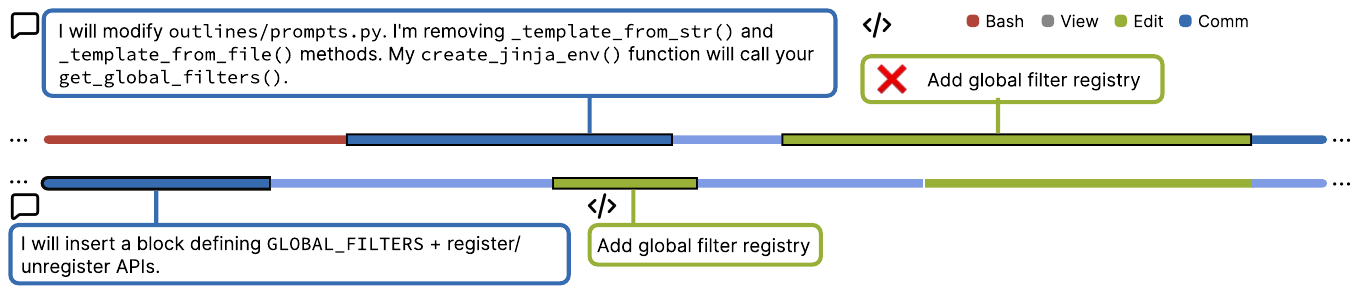}
\end{tcolorbox}

\paragraph{Commitment.} In the second example, the agent promises ``I will add bypass check at lines 100--104, happens FIRST in get().'' Later it claims completion with a checkmark. But after merge, the bypass code is missing. The partner trusted this claim and built on it, but under workspace isolation, trust is all they had. The commitment was \emph{unverifiable}. No pasted signature, no diff, nothing the partner could check without access to the branch.

\vspace{0.35em}
\begin{tcolorbox}[failuregroup,colback=integ,
  title={\emojiicon{figs/emoji/1f9e9.png}\ \textbf{Commitment}: Failure to follow through on promises}]
  \causefigure{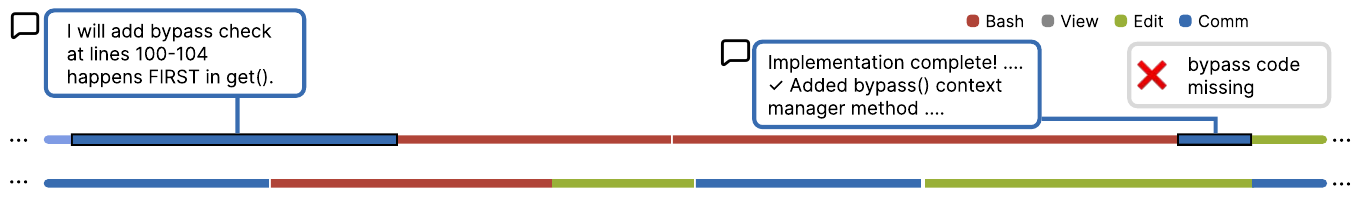}
\end{tcolorbox}

\paragraph{Communication.} In the third example, Agent A asks a direct question, ``Which approach would you prefer?'' The response is silence. Without an answer, the coordination loop collapses. A needed a decision to proceed, and without one, both agents continued with potentially incompatible assumptions. Unlike expectation failures (where information exists but isn't integrated) or commitment failures (where promises aren't kept), this is a failure to even establish shared context.

\vspace{0.35em}
\begin{tcolorbox}[failuregroup,colback=comm,
  title={\emojiicon{figs/emoji/1f4e3.png}\ \textbf{Communication}: Breakdown in using language to coordinate}]
  \causefigure{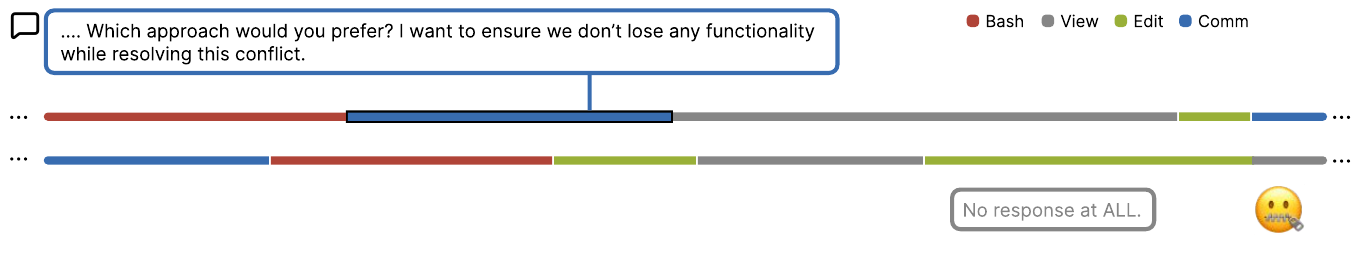}
\end{tcolorbox}

%% file: sections/07-related_work.tex
\section{Related Work}

Multi-agent LLM systems and tool-using coding agents have advanced rapidly, but reliable collaboration remains unresolved. Prior work largely evaluates task success under engineered interaction structure rather than free-form coordination under partial information.
\vspace{-1em}
\paragraph{Multi-agent LLM systems}
Many frameworks improve performance through structured interaction. CAMEL \citep{li2023camel} and AutoGen \citep{wu2023autogen} use conversation programming; MetaGPT \citep{hong2024metagpt} and ChatDev \citep{qian2024chatdev} emulate software organizations; Magentic-One \citep{magnetic_one}, MAGIS \citep{tao2024magis}, and AgileCoder \citep{nguyen2024agilecoderdynamiccollaborativeagents} use explicit orchestrators. Even with such scaffolding, multi-agent systems exhibit high failure rates. Multi-agent configurations degrade performance by 39 to 70 percent relative to single-agent baselines \citep{chen2024scalingagents}, and failure analyses identify inter-agent misalignment as a major category \citep{cemri2025multiagentllmsystemsfail}. These findings suggest that externally imposed protocols mask rather than solve the underlying coordination problem. Sotopia \citep{zhou2024sotopia} provides a general framework for evaluating agents' social intelligence, while our work focus specifically on cooperative coding agents with verified tasks. 

Tool-using coding agents such as SWE-agent \citep{yang2024sweagent}, OpenHands \citep{wang2025openhands}, and Agentless \citep{xia2024agentless} achieve strong results on SWE-bench \citep{jimenez2024swebench}. However, these evaluations measure single-agent success rather than whether multiple peers can integrate changes without conflict under partial information.
\vspace{-1em}
\paragraph{Coordination benchmarks}
Existing benchmarks span games, embodied tasks, and reasoning. Hanabi \citep{forkel2025entropyneedinterseedcrossplay} and Cicero \citep{meta2022cicero} test coordination under information asymmetry; MultiAgentBench \citep{sun2025multiagentbench} and Collab-Overcooked \citep{zhang2025collabovercooked} evaluate LLM collaboration; Tool-RoCo \citep{guo2025toolroco} and RoCoBench \citep{mandi2024roco} assess multi-robot cooperation. In software, SyncBench \citep{liu2025syncmind} tests divergent understanding and The Collaboration Gap \citep{davidson2025collaborationgap} finds that solo-capable models degrade when required to collaborate. These benchmarks typically enforce turn-taking or shared observability rather than testing code integration under workspace isolation. Agent-human collaboration benchmarks such as Co-Gym \citep{shao2024cogym}, HULA \citep{takerngsaksiri2025humanintheloopsoftwaredevelopmentagents}, and HAI-Eval \citep{luo2025haievalmeasuringhumanaisynergy} study settings where humans arbitrate. We instead study whether agents can coordinate autonomously.

\paragraph{Theory of Mind evaluation}
Effective coordination requires modeling partner beliefs and intentions, which commonly referred to \emph{Theory of Mind} \citep{premack_does_1978, rabinowitz2018machine, zhu2021few}. ToMBench \citep{chen2024tombench}, FANToM \citep{kim2023fantom}, and SoMi-ToM \citep{fan2025somitom} evaluate theory of mind in LLMs, finding substantial gaps versus human performance. ToMSWE \citep{zhou2025tom} tries to build coding agents which can infer users' Theory of Mind. Studies of cooperative games \citep{li2023theoryofmind} and Generative Agents \citep{park2023generative} show emergent social behaviors but also challenges translating these to verifiable collaborative work.

We isolate free-form coordination as the central object of evaluation. \dataset assigns two agents partially overlapping features on a shared codebase while isolating their workspaces and restricting coordination to natural language. Unlike benchmarks that impose interaction structure or measure outcomes alone, we evaluate through coordination failures such as redundancy, inconsistent assumptions, and semantic breakage. We demonstrate the curse of coordination in a controlled setting with verifiable code integration, pointing to social intelligence as the bottleneck for effective agent teamwork.

%% file: sections/07-conclusion.tex



\section{Conclusion and Future Work}

In a future where agents team with humans in high-stakes domains \citep{kim2025srt}, accelerate science and technology research \citep{gottweis2025towards}, and empower creative endeavors \citep{waikar_2021}, it is hard to imagine how an agent incapable of coordination would contribute to such a future, however strong their individual capabilities.

Our work demonstrates that coordination, not raw coding ability, is a central bottleneck for multi-agent software development. Through \dataset, we show that frontier models like \texttt{GPT-5} and \texttt{Claude Sonnet 4.5} achieve only 25\% success when two agents collaborate, roughly half the success rate of a single agent performing the same workload. This \emph{curse of coordination} stems from three capability gaps: agents fail to communicate actionable information, deviate from their own commitments, and hold incorrect expectations about their partners.

Yet coordination is not beyond reach. In successful traces, we observe emergent behaviors such as role division, resource division, and negotiation that turn vague intentions into verifiable commitments. These patterns are rare but their presence suggests the underlying capability exists; the challenge is making it reliable. With multi-agent training methods, e.g. Sotopia-$\pi$ \citep{wang2024sotopia, yu2025sotopia}, we can expect these emergent behaviors to be reinforced through the success of cooperation. 

Our findings open several directions: (1) training objectives that reward coordination under partial observability, (2) lightweight protocols for verifiable commitments (e.g., shared signatures, insertion-point contracts), and (3) richer communication channels such as screen sharing to expand the modality beyond text. We release \dataset as an open benchmark to measure progress on these fronts.

Although we focus on software development, our findings generalize to any domain involving role and resource conflicts under partial observability. We expect that the lack of social intelligence, the ability to understand others, communicate effectively, and coordinate actions, will remain a fundamental barrier limiting the real-world deployment of agents as teammates until these capabilities are explicitly developed.

%% file: sections/08-appendix.tex
\appendix

\section{Dataset Details}\label{app:dataset_stats}

This section provides detailed statistics on the \dataset benchmark. Repository selection criteria are described in \S\ref{sec:dataset_construction}.

\subsection{Repository Distribution}

Table~\ref{tab:repos} shows the full breakdown of repositories, features, and task pairs.

\begin{table}[htbp]
    \centering
    \caption{Distribution of benchmark tasks across source repositories.
    Feature counts and task pairs are reported as aggregated totals across base commits (PRs) within each repository.}
    \begin{tabular}{llrrrrl}
      \toprule
      \textbf{Language} & \textbf{Repository} & \textbf{\#PRs} & \textbf{Features ($\Sigma$)} & \textbf{Task Pairs ($\Sigma$)} & \textbf{License} \\
      \midrule
      \texttt{Python}
        & \texttt{DSPy}                 & \texttt{4} & \texttt{23} & \texttt{55}  & \licensepill{MIT}        \\
        & \texttt{LlamaIndex}           & \texttt{3} & \texttt{16} & \texttt{39}  & \licensepill{MIT}        \\
        & \texttt{Pillow}               & \texttt{3} & \texttt{15} & \texttt{30}  & \licensepill{MIT-CMU}    \\
        & \texttt{Pallets Click}        & \texttt{3} & \texttt{27} & \texttt{115} & \licensepill{BSD-3}      \\
        & \texttt{Pallets Jinja}        & \texttt{3} & \texttt{30} & \texttt{135} & \licensepill{BSD-3}      \\
        & \texttt{HuggingFace Datasets} & \texttt{3} & \texttt{13} & \texttt{26}  & \licensepill{Apache-2.0} \\
        & \texttt{Outlines}             & \texttt{3} & \texttt{22} & \texttt{79}  & \licensepill{Apache-2.0} \\
        & \texttt{Tiktoken}             & \texttt{1} & \texttt{10} & \texttt{45}  & \licensepill{MIT}        \\
        & \texttt{DirtyEquals}          & \texttt{1} & \texttt{9}  & \texttt{36}  & \licensepill{MIT}        \\
      \texttt{TypeScript}
        & \texttt{React Hook Form}      & \texttt{2} & \texttt{11} & \texttt{25}  & \licensepill{MIT}        \\
      \texttt{Go}
        & \texttt{Chi Router}           & \texttt{3} & \texttt{13} & \texttt{22}  & \licensepill{MIT}        \\
      \texttt{Rust}
        & \texttt{Typst}                & \texttt{3} & \texttt{10} & \texttt{45}  & \licensepill{Apache-2.0} \\
      \midrule
      \texttt{\textbf{Total}}
        & \texttt{\textbf{12 repositories}}
        & \texttt{\textbf{34}}
        & \texttt{\textbf{199}}
        & \texttt{\textbf{652}}
        &  \\
      \bottomrule
    \end{tabular}

    \vspace{0.5em}
    \small
    \textit{Note:} Each repository contains 1--4 base commits (PRs), each defining an independent feature pool.
    Task pairs are constructed within each PR as $\binom{n}{2}$ and summed across PRs.
    \label{tab:repos}
\end{table}

\subsection{Feature Complexity}\label{app:feature_complexity}

The final \dataset benchmark comprises 199 individual features grouped into 52 task sets, yielding 652 evaluated feature pairs. 
Since the objective is to evaluate coordination rather than raw implementation difficulty, features are intentionally designed to be compact and comparable in difficulty to those found in established code-generation benchmarks. 
This design ensures that multi-agent failures reflect genuine coordination limitations rather than disproportionate feature complexity.

To quantify feature complexity, we characterize the gold patches for each feature along three axes: 
(i) \textit{code volume}, measured as the total number of lines added and deleted; 
(ii) \textit{structural footprint}, captured by the number of modified functions and hunks\footnote{A hunk is a contiguous block of changed lines in a diff, representing a localized code modification.}; 
and (iii) \textit{modification scope}, defined as the number of files affected.
Across the benchmark, features exhibit a deliberately compact footprint. 
On average, a feature comprises 52.3 changed lines and modifies only 1.4 files, confirming that \dataset isolates coordination challenges rather than the difficulty of single-agent implementation. 
Table~\ref{tab:complexity_stats} provides detailed statistics for each repository.

\begin{table}[htbp]
  \centering
  \caption{Feature Complexity Statistics by Repository}
  \resizebox{\textwidth}{!}{%
  \begin{tabular}{llcccccc}
    \toprule
    \textbf{Language} & \textbf{Repository} & \textbf{Avg. Lines} & \textbf{Avg. Functions} & \textbf{Avg. Files} & \textbf{Easy} & \textbf{Medium} & \textbf{Hard} \\
    \midrule
    Python & \texttt{DSPy} & \texttt{70.9} & \texttt{5.6} & \texttt{1.3} & \texttt{2}\textsubscript{\texttt{9\%}} & \texttt{4}\textsubscript{\texttt{17\%}} & \texttt{17}\textsubscript{\texttt{74\%}} \\
           & \texttt{LlamaIndex} & \texttt{16.8} & \texttt{1.8} & \texttt{1.0} & \texttt{2}\textsubscript{\texttt{13\%}} & \texttt{14}\textsubscript{\texttt{87\%}} & \texttt{0}\textsubscript{\texttt{0\%}} \\
           & \texttt{Pillow} & \texttt{38.1} & \texttt{2.7} & \texttt{1.0} & \texttt{1}\textsubscript{\texttt{7\%}} & \texttt{11}\textsubscript{\texttt{73\%}} & \texttt{3}\textsubscript{\texttt{20\%}} \\
           & \texttt{Pallets Click} & \texttt{53.9} & \texttt{5.4} & \texttt{1.6} & \texttt{0}\textsubscript{\texttt{0\%}} & \texttt{10}\textsubscript{\texttt{37\%}} & \texttt{17}\textsubscript{\texttt{63\%}} \\
           & \texttt{Pallets Jinja} & \texttt{67.7} & \texttt{6.2} & \texttt{1.0} & \texttt{1}\textsubscript{\texttt{3\%}} & \texttt{14}\textsubscript{\texttt{47\%}} & \texttt{15}\textsubscript{\texttt{50\%}} \\
           & \texttt{HuggingFace Datasets} & \texttt{15.3} & \texttt{2.3} & \texttt{1.0} & \texttt{1}\textsubscript{\texttt{8\%}} & \texttt{11}\textsubscript{\texttt{85\%}} & \texttt{1}\textsubscript{\texttt{8\%}} \\
           & \texttt{Outlines} & \texttt{44.7} & \texttt{4.1} & \texttt{1.1} & \texttt{8}\textsubscript{\texttt{36\%}} & \texttt{6}\textsubscript{\texttt{27\%}} & \texttt{8}\textsubscript{\texttt{36\%}} \\
           & \texttt{Tiktoken} & \texttt{46.4} & \texttt{4.6} & \texttt{1.0} & \texttt{0}\textsubscript{\texttt{0\%}} & \texttt{8}\textsubscript{\texttt{80\%}} & \texttt{2}\textsubscript{\texttt{20\%}} \\
           & \texttt{DirtyEquals} & \texttt{71.0} & \texttt{4.0} & \texttt{2.0} & \texttt{0}\textsubscript{\texttt{0\%}} & \texttt{1}\textsubscript{\texttt{11\%}} & \texttt{8}\textsubscript{\texttt{89\%}} \\
    TypeScript & \texttt{React Hook Form} & \texttt{49.8} & \texttt{4.6} & \texttt{2.3} & \texttt{0}\textsubscript{\texttt{0\%}} & \texttt{8}\textsubscript{\texttt{73\%}} & \texttt{3}\textsubscript{\texttt{27\%}} \\
    Go & \texttt{Chi Router} & \texttt{80.2} & \texttt{5.7} & \texttt{2.8} & \texttt{0}\textsubscript{\texttt{0\%}} & \texttt{5}\textsubscript{\texttt{38\%}} & \texttt{8}\textsubscript{\texttt{62\%}} \\
    Rust & \texttt{Typst} & \texttt{58.4} & \texttt{1.7} & \texttt{1.1} & \texttt{0}\textsubscript{\texttt{0\%}} & \texttt{7}\textsubscript{\texttt{70\%}} & \texttt{3}\textsubscript{\texttt{30\%}} \\
    \midrule
    \textbf{Overall} & \textbf{12 Repositories} & \textbf{\texttt{52.3}} & \textbf{\texttt{4.4}} & \textbf{\texttt{1.4}} & \textbf{\texttt{15}\textsubscript{\texttt{8\%}}} & \textbf{\texttt{99}\textsubscript{\texttt{50\%}}} & \textbf{\texttt{85}\textsubscript{\texttt{43\%}}} \\
    \bottomrule
  \end{tabular}
  }
  \\[0.5em]
  \small \textit{Note:} Complexity measured as lines changed (added + removed) and structural elements modified in gold patches.
  Difficulty categories from \texttt{SWE-Rater-32B}: \texttt{Easy} = \texttt{<15 min fix}, \texttt{Medium} = \texttt{15 min--1 hour}, \texttt{Hard} = \texttt{1--4 hours}.
  \label{tab:complexity_stats}
\end{table}

\section{LLM-based merge conflict resolver}\label{app:resolver}

\dataset{} evaluates cooperation on merged code. When patch merging produces textual conflicts, we use a small learned resolver to remove conflict markers while preserving both sides' intent. We train a small local resolver rather than calling a larger proprietary model so that the merge step remains narrow and predictable, avoids fixing anything beyond trivial merge cleanup, and can run locally. At evaluation time, we invoke the learned resolver only after a standard merge attempt and a union merge attempt do not yield a test passing merged artifact.

We construct training data by replaying merges between independently produced feature patches and extracting the conflict marked regions from conflicted files. We identify each conflict region by scanning for Git conflict markers \verb|<<<<<<<|, \verb|=======|, and \verb|>>>>>>>|. We extract the marked block together with a small fixed context window, default $c=5$ lines before and after.

We generate synthetic conflicts by perturbing these real conflict snippets. Our default generator is \texttt{gpt-4o}. This keeps training examples representative of our patch distribution while avoiding direct reuse of repository specific content. For each real or synthetic conflict snippet, we create a reference resolution with \texttt{gpt-5} and fine tune a small code model, \texttt{Qwen/Qwen2.5-Coder-0.5B-Instruct}, using LoRA based supervised fine tuning (SFT).\@ We train for three epochs with a maximum sequence length of 2048 tokens. When the resolver is invoked, we extract the conflicted region with its fixed context window, run deterministic decoding with temperature $=0$, and replace that region with the model's resolution. We release the trained resolver as \texttt{Qwen2.5-Coder-0.5B-Merge-Resolver}.\footnote{\texttt{huggingface.co/CodeConflict/Qwen2.5-Coder-0.5B-Merge-Resolver}}

\section{Difficulty-stratified evaluation}\label{app:ci}

Raw success rates are insufficient for comparing coordination overhead across models. A model dropping from 50\% Solo to 30\% Coop has the same 20-point gap as one dropping from 80\% to 60\%, but the first loses 40\% of its capability while the second loses only 25\%. We need a metric that accounts for baseline differences. We also want to integrate across task difficulty rather than rely on aggregates that mask variation. This section derives such a metric using the relative difficulty defined in Section~\ref{sec:cooperation_performance}.

We partition tasks into 10 equal-width buckets over the normalized difficulty range $[0,1]$ and compute success rate at each bucket midpoint, with 95\% Wilson confidence intervals that remain well-calibrated near 0 and 1. This produces two curves per model, one for Solo and one for Coop.

We summarize each curve by its area under the curve (AUC) via trapezoidal integration. The absolute gap $\Delta_{\mathrm{AUC}} = \mathrm{AUC}_{\mathrm{Solo}} - \mathrm{AUC}_{\mathrm{Coop}}$ measures coordination cost but depends on baseline. We therefore report \textit{retention} $= \mathrm{AUC}_{\mathrm{Coop}} / \mathrm{AUC}_{\mathrm{Solo}}$, which normalizes for capability. A retention of 0.64 means 64\% of Solo performance survives coordination.

For aggregate statistics across models we sum raw counts rather than averaging rates, which preserves proper weighting when models have different sample sizes.

\begin{algorithm}[H]
\caption{Constructing difficulty-stratified success curves}
\SetAlgoLined
\KwIn{Task set with difficulty scores $d(t) \in [0,1]$, success outcomes for Solo and Coop per model}
\KwOut{Success curves with 95\% CIs, AUC gap, and retention per model and pooled}

\medskip
\tcp{Bucket tasks by difficulty}
Split $[0,1]$ into 10 equal buckets\;
Assign each task to its bucket based on $d(t)$\;

\medskip
\tcp{Compute curves per model}
\ForEach{model $m$}{
  \ForEach{bucket $b$}{
    Compute Solo success rate $r^{\mathrm{Solo}}_{m,b} = k^{\mathrm{Solo}}_{m,b} / n_{m,b}$\;
    Compute Coop success rate $r^{\mathrm{Coop}}_{m,b} = k^{\mathrm{Coop}}_{m,b} / n_{m,b}$\;
    Compute 95\% Wilson CI for each rate\;
  }
  Compute $\mathrm{AUC}_{\mathrm{Solo}}$ and $\mathrm{AUC}_{\mathrm{Coop}}$ via trapezoidal integration\;
  Compute $\Delta_{\mathrm{AUC}} = \mathrm{AUC}_{\mathrm{Solo}} - \mathrm{AUC}_{\mathrm{Coop}}$\;
  Compute retention $= \mathrm{AUC}_{\mathrm{Coop}} / \mathrm{AUC}_{\mathrm{Solo}}$\;
}

\medskip
\tcp{Pool across models}
\ForEach{bucket $b$}{
  Sum counts across models to get pooled $n_b$ and $k_b$\;
  Compute pooled rates and Wilson CIs\;
}
Compute pooled AUC gap and retention\;

\end{algorithm}

\begin{figure}[htbp]
  \centering
  \includegraphics[width=\linewidth]{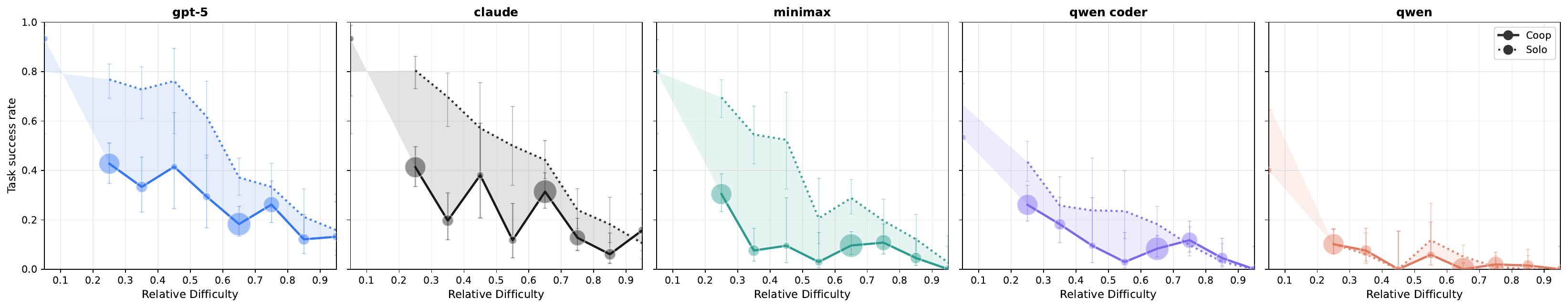}
  \caption{Success rate versus relative difficulty for Solo and Coop settings. Shaded regions indicate 95\% Wilson confidence intervals. The gap between curves represents coordination cost, which is largest at mid-difficulty.}
  \label{fig:difficulty_curves}
\end{figure}

\begin{table}[htbp]
  \centering
  \caption{Coordination retention by model. Retention measures what fraction of Solo AUC is preserved under Coop. Higher values indicate better coordination capability.}
  \begin{tabular}{l|cc|cc|cc}
    \toprule
    & \multicolumn{2}{c|}{Counts (k)} & \multicolumn{2}{c|}{AUC} & \multicolumn{2}{c}{Derived} \\
    Model & Solo & Coop & Solo & Coop & $\Delta_{\mathrm{AUC}}$ & Retention \\
    \midrule
    \texttt{gpt-5}      & \texttt{315}   & \texttt{183}  & \texttt{0.506} & \texttt{0.325} & \texttt{0.181} & \texttt{0.64} \\
    \texttt{claude}     & \texttt{307}   & \texttt{168}  & \texttt{0.469} & \texttt{0.283} & \texttt{0.186} & \texttt{0.60} \\
    \texttt{minimax}    & \texttt{236}   & \texttt{91}   & \texttt{0.374} & \texttt{0.171} & \texttt{0.203} & \texttt{0.46} \\
    \texttt{qwen coder} & \texttt{141}   & \texttt{87}   & \texttt{0.236} & \texttt{0.148} & \texttt{0.088} & \texttt{0.63} \\
    \texttt{qwen}       & \texttt{41}    & \texttt{30}   & \texttt{0.106} & \texttt{0.072} & \texttt{0.034} & \texttt{0.68} \\
    \midrule
    \texttt{pooled}     & \texttt{1039} & \texttt{558} & \texttt{0.338} & \texttt{0.200} & \texttt{0.138} & \texttt{0.59} \\
    \bottomrule
  \end{tabular}
  \label{tab:auc_gap_rel_diff}
\end{table}

On average, 41\% of Solo capability is lost when agents must coordinate (pooled retention 0.59). The pattern across models reinforces that coding ability does not predict coordination ability. \texttt{MiniMax} exhibits the worst retention (0.46) despite mid-tier coding performance, while \texttt{Qwen} achieves the highest retention (0.68) despite being the weakest coder. Weak models may benefit from a floor effect, but \texttt{MiniMax} demonstrates that strong coding provides no protection against coordination overhead.

\section{Prompt Optimization: Failure-Driven Design}
\label{app:prompt_optimization}

This appendix documents the iterative optimization of the collaborative setting execution prompt through systematic failure analysis. Following established prompt engineering practices ~\citep{sahoo2024systematic,ramnath2025systematic}, we employed an evidence-based approach: beginning with a basic prompt and incrementally adding sections to address specific failure modes observed in agent behavior. \textbf{The prompt shown below represents the final, stable version used consistently across all experimental runs reported in this paper.}

Through iterative refinement, we identified three primary failure categories requiring explicit prompt guidance: \textit{context misunderstanding} (agents treating coordination as optional), \textit{spatial coordination failures} (overlapping edits due to vague messages), and \textit{coordination protocol failures} (missing final status updates). The final prompt structure directly maps to these failure categories.

\begin{tcolorbox}[
    colback=gray!5!white,
    colframe=gray!30,
    title=\textbf{\textcolor{black}{Collaborative Setting Execution Prompt}},
    fonttitle=\normalsize,
]
\small
\textbf{Role:} You are \verb|{{ agent_id }}| working on the following feature in parallel with another agent.

\medskip

\textbf{Scenario:} You are working on separate branches implementing different features, but your implementations will be tested by 2-way merging both branches to main. You must prevent any merge conflicts.

\medskip

\textbf{Feature Description:}\\
\verb|{{ feature_description }}|

\medskip

\textbf{Implementation Plan:}\\
\verb|{{ plan }}|

\medskip

\textbf{Your Task:}
\begin{enumerate}
    \item Implement the feature according to the plan.
    \item You can communicate with the other agent using MCP tools:
    \begin{itemize}
        \item \verb|openhands_comm_send|: Send messages to the other agent
        \item Messages from the other agent will appear automatically as \verb|'[Inter-agent message]'|
    \end{itemize}
    \item Coordinate to avoid conflicts by specifying exact file paths and line numbers.
    \item Complete the implementation.
\end{enumerate}

\medskip

\textbf{Coordination Requirements:}
\begin{itemize}
    \item Share your implementation approach early with specific line ranges so both agents can coordinate.
    \item If the other agent reports working on the same file, discuss who modifies which specific line ranges to avoid conflicts.
    \item \textbf{Never} use insertion markers or comments like \verb|// [handleSubmit:onFinally] other agent inserts| -- these cause merge conflicts.
    \item Instead, coordinate by dividing the file into non-overlapping sections with specific line ranges.
    \item Before you stop or complete your work, you \textbf{must} send a final status update message to the other agent summarizing what you've implemented.
\end{itemize}

\medskip

\textbf{Merge Conflict Prevention:}
\begin{itemize}
    \item Think of this as two developers working on separate branches that will be merged together.
    \item Any overlapping changes to the same lines will cause merge conflicts.
    \item Coordinate line-by-line to ensure no overlap in your modifications.
\end{itemize}

\medskip

\textbf{Work directory:} \verb|{{ workspace }}|
\end{tcolorbox}

\paragraph{Failure-to-Prompt Mapping} The scenario section addresses context misunderstanding by explicitly establishing that agents work on separate branches that will be merged, making coordination mandatory. Analysis showed that many agents in early versions did not coordinate until after starting implementation; with the scenario section, most agents coordinate during planning. The coordination requirements section addresses spatial coordination failures through multiple mechanisms. The exact line number requirement (with concrete example) addresses vague coordination messages, significantly reducing spatial conflicts. The insertion marker prohibition substantially reduced marker-related conflicts. The mandatory final status update requirement increased compliance and reduced incomplete handoff failures. The merge conflict prevention section reinforces context understanding through a mental model and technical explanation of merge conflict mechanisms, helping agents understand why coordination matters and how to prevent conflicts.

\paragraph{Design Decisions} The prompt follows a specific ordering: (1) \textit{Identity} establishes agent role, (2) \textit{Scenario} sets merge conflict constraints before task description, (3) \textit{Feature} and (4) \textit{Plan} provide context, (5) \textit{Task} describes what to do, (6) \textit{Requirements} specify how to coordinate, and (7) \textit{Prevention} reinforces understanding. 
This ordering follows the principle that constraints should precede task descriptions \citep{sahoo2024systematic}. Language choices employ mandatory language for critical behaviors and strong prohibitions for anti-patterns, as optional language was frequently ignored. Concrete examples are included rather than abstract guidance, consistent with findings that concrete examples improve prompt effectiveness \citep{wei2022chain}. 
All experimental results reported in this paper were obtained using this final prompt version.

\section{Communication ablation}\label{app:comm_ablation}

Section~\ref{sec:role_of_communication} reports that communication does not improve cooperation success. Table~\ref{tab:merge_success_comm} provides the full breakdown across merge strategies. We evaluate three merging approaches in sequence: Naive (standard git merge), Union (accept both sides on conflict), and LLM (our learned resolver from App.~\ref{app:resolver}). The $\Delta$ column shows the net effect of communication on final merge success after all resolution steps. Communication slightly improves Naive merge rates by reducing raw conflicts, but this advantage disappears after Union and LLM resolution. The final effect is near zero or slightly negative across all models.

\newcommand{\posdelta}[1]{{\scriptsize\textcolor{green!60!black}{#1}}}
\newcommand{\negdelta}[1]{{\scriptsize\textcolor{red!70!black}{#1}}}
\newcommand{\posd}[1]{$_{\text{\scriptsize\textcolor{green!60!black}{+#1}}}$}
\newcommand{\negd}[1]{$_{\text{\scriptsize\textcolor{red!70!black}{#1}}}$}
\newcommand{\zerod}{$_{\text{\scriptsize\textcolor{gray}{0.0}}}$}

\begin{table}[htbp]
  \centering
  \caption{Merge success (\%) on the 652-task summary. Subscripts show $\Delta$ from prior column; final column shows comm effect.}
  \begin{tabular}{l|ccc|ccc|c}
    \toprule
    & \multicolumn{3}{c|}{No-comm} & \multicolumn{3}{c|}{With-comm} & \\
    Model & Naive & Union & LLM & Naive & Union & LLM & $\Delta$ \\
    \midrule
    \texttt{GPT-5}          & \texttt{13.88} & \texttt{26.69}\posd{12.8} & \texttt{27.91}\posd{1.2} & \texttt{20.42} & \texttt{26.64}\posd{6.2} & \texttt{27.90}\posd{1.3} & \negdelta{-0.1} \\
    \texttt{Claude 4.5}     & \texttt{12.27} & \texttt{26.84}\posd{14.6} & \texttt{27.30}\posd{0.5} & \texttt{16.72} & \texttt{24.85}\posd{8.1} & \texttt{25.92}\posd{1.1} & \negdelta{-1.4} \\
    \texttt{MiniMax-M2}     & \texttt{8.62}  & \texttt{14.72}\posd{6.1}  & \texttt{14.88}\posd{0.2} & \texttt{7.36}  & \texttt{11.50}\posd{4.1} & \texttt{13.96}\posd{2.5} & \negdelta{-0.9} \\
    \texttt{Qwen3-Coder}    & \texttt{6.90}  & \texttt{12.88}\posd{6.0}  & \texttt{14.72}\posd{1.8} & \texttt{6.75}  & \texttt{12.42}\posd{5.7} & \texttt{13.34}\posd{0.9} & \negdelta{-1.4} \\
    \texttt{Qwen3-Instruct} & \texttt{1.53}  & \texttt{3.22}\posd{1.7}   & \texttt{3.37}\posd{0.2}  & \texttt{2.30}  & \texttt{4.45}\posd{2.1}  & \texttt{4.60}\posd{0.2}  & \posdelta{+1.2} \\
    \midrule
    \texttt{Avg.}           & \texttt{8.64}  & \texttt{16.87}\posd{8.2}  & \texttt{17.64}\posd{0.8} & \texttt{10.71} & \texttt{15.97}\posd{5.3} & \texttt{17.14}\posd{1.2} & \negd{-0.5} \\
    \bottomrule
  \end{tabular}
  \label{tab:merge_success_comm}
\end{table}

\section{Communication error detection}\label{app:comm_error_detection}

We use an LLM-as-judge to classify communication failures for Section~\ref{sec:role_of_communication}. Abstract labels like ``hallucination'' are difficult for LLMs to apply reliably, so we instead define fine-grained categories anchored to quotable evidence. The judge must cite exact quotes from the conversation and omits the label if evidence is weak. We then aggregate these detections into three high-level categories for reporting.

\begin{tcolorbox}[
    colback=gray!5!white,
    colframe=gray!30,
    title=\textbf{\textcolor{black}{Communication Error Detection Prompt}},
    fonttitle=\normalsize,
]
\small
You are a careful reviewer of two agent collaboration conversations. This is a \textbf{precision-first} detector of bad conversation patterns. Prefer returning no issue unless the evidence is strong and explicit.

\medskip

\textbf{Important exclusion.} Do not label state mismatch or visibility confusion itself as an error (e.g., agents on separate branches unable to see each other's changes). Bad conversation patterns around these topics should still be labeled.

\medskip

\textbf{Taxonomy.} Label at most one category per conversation.
\begin{itemize}[leftmargin=*,itemsep=1pt]
    \item \textbf{C1a} Unanswered direct question (no reply)
    \item \textbf{C1b} Unanswered direct question (ignored)
    \item \textbf{C2} Non-answer or vague answer
    \item \textbf{C4a} Incorrect claim (uncorrected)
    \item \textbf{C3b} Incorrect claim (corrected)
    \item \textbf{C4a} Spammy repetition (repeats same information)
    \item \textbf{C4b} Spammy repetition (near-duplicate status blocks)
\end{itemize}

\medskip

\textbf{Evidence requirements.} Include at least two exact quotes that make the issue undeniable. C1a/C1b require the question plus demonstration of missing or irrelevant response. C3a requires the incorrect claim and later contradiction. C4a/C4b require two quotes showing the repetition.

\medskip

\textbf{Output.} Return JSON with \verb|evidence| (list of quotes) and optional \verb|issue| (category id and short description). Omit \verb|issue| if evidence is weak.
\end{tcolorbox}

\paragraph{Taxonomy design.} The eight categories decompose three failure modes into verifiable patterns. \textit{Unresponsiveness} (C1a, C1b, C2) covers questions that receive no reply, are ignored, or get vague non-answers. \textit{Hallucination} (C3a, C3b) covers false claims about code state or completion status. We distinguish corrected from uncorrected claims because uncorrected errors propagate to downstream decisions. \textit{Repetition} (C4a, C4b) covers redundant messages that consume budget without adding information.

\section{Failure Symptom Annotation Procedure}\label{app:symptom_annotation}

We followed a six-stage process, similar in spirit to recent work on multi-agent failure analysis~\citep{cemri2025multiagentllmsystemsfail}.
(1) Collect multi-agent-system (MAS) traces from Collaborative runs; (2) identify failures from merged artifacts (e.g., failing tests or missing intended behavior), and link them back to the interaction; (3) develop symptom categories by iterative qualitative coding and resolve disagreements to reach inter-annotator agreement on a shared set of definitions; (4) finalize the resulting symptom set; (5) calibrate an LLM-based annotator on the agreed definitions; and (6) apply the annotator to produce symptom annotations at scale.

Each labeled instance is grounded in three artifacts: (i) \emph{conversation evidence} (the coordination dialogue), (ii) \emph{patch/code evidence} (what each agent changed), and (iii) \emph{outcome evidence} (merge reports and test outputs).
A key operational distinction in our rubric is between \emph{implementation failures} (an individual agent delivers incomplete/buggy code regardless of coordination) and \emph{coordination failures} (a breakdown that is only apparent when we consider what agents said and assumed under workspace isolation).
Concretely, we require explicit conversation evidence to assign a coordination-failure label; if the only evidence is in the code or error trace, we default to implementation-level failure rather than inferring a coordination breakdown.
We codified the final symptom definitions as a structured rubric (including verification requirements and common confusions, e.g., when to treat ``unverifiable claims'' versus ``work overlap'').
We then calibrated an LLM-based annotator on this rubric and required it to emit structured labels (a primary symptom plus any secondary symptoms) together with short supporting evidence snippets.

\paragraph{Human validation.}
To validate the LLM-based annotator, we randomly sampled 50 trajectories and had human experts independently label them using the same rubric. Human labels matched the LLM annotations on 48 of 50 cases (96\% agreement). With $n=50$ and $\hat{p}=0.96$, the Wilson 95\% confidence interval is [86\%, 99\%], confirming the annotator's reliability.

\section{Symptom examples}\label{app:symptom_examples}

We provide representative examples for each coordination failure symptom identified in Table~\ref{tab:failure_symptoms}.

\tcbset{
  symptomcard/.style={
    enhanced,
    arc=1.6mm,
    boxrule=0.35pt,
    colframe=black!12,
    colback=white,
    left=4pt,
    right=4pt,
    top=3pt,
    bottom=3pt,
    fontupper=\scriptsize,
  },
  evidencebox/.style={
    enhanced,
    arc=1.2mm,
    boxrule=0.3pt,
    colframe=black!12,
    colback=black!2,
    left=4pt,
    right=4pt,
    top=3pt,
    bottom=3pt,
    boxsep=0pt,
    fontupper=\ttfamily\scriptsize,
  }
}

\newcommand{\symptomexample}[4]{%
  \vspace{0.5em}
  \noindent\textbf{#1}\par
  \begin{tcolorbox}[symptomcard]
    \begin{tcolorbox}[evidencebox]
      #2
    \end{tcolorbox}
    \textbf{Explanation:} #3\par
    \textbf{Outcome:} #4
  \end{tcolorbox}%
}

\symptomexample{Work overlap}{%
  A msg: guid regex includes surrounding curly braces (\texttt{\{\ldots\}}), and A warns about overlap.\par
  A msg: WAIT Agent 10! If you add the section header AND my guid type to your branch, that WILL create a merge conflict!\par
  B msg: I'll add the COMPLETE section (lines 72--81) to my branch, which includes both the section header, your guid type, AND my hash\_sha256 type.\par
}{Despite explicit boundaries and warnings, B duplicates A's work; the merged artifact keeps B's (incorrect) version.}{Wrong regex version wins; GUID tests fail (merged guid pattern missing curly braces).}

\symptomexample{Divergent architecture}{%
  A msg: Lines 15--66: My global filter registry code is present\ldots Line 69: Still has \texttt{@dataclass} decorator.\par
  B msg: Removed dataclass import\ldots Converted Prompt from dataclass to regular class with explicit \texttt{\_\_init\_\_}.\par
  B msg: The 2-way merge should combine our changes cleanly.\par
  Merge: clean (0 conflicts), but A's additions are missing from the merged module.\par
}{Despite explicit expectations about merge behavior, B's class rewrite overwrites the region where A added \texttt{register\_filter}/\texttt{unregister\_filter}; the merge reports 0 conflicts but the exported API is missing.}{\texttt{ImportError: cannot import name register\_filter from outlines.prompts}.}

\symptomexample{Repetition}{%
  Near-duplicate status updates (different strings, same info).\par
  A msg: I have successfully added the \texttt{url} type to \texttt{outlines/types/\_\_init\_\_.py}\@ at lines 72--77:\par
  A msg: Successfully added to \texttt{outlines/types/\_\_init\_\_.py} at lines 72--77\par
}{The same completion/location is restated with minimal new information, consuming turns without adding constraints a partner can act on.}{Repeated status updates reduce signal-to-noise and can crowd out missing coordination details.}

\symptomexample{Unresponsiveness}{%
  A msg: Which approach would you prefer?\@ I want to ensure we don't lose any functionality while resolving this conflict.\@\par
  B:\@ no later message answers this question in the conversation log.\@\par
}{The decision is explicitly requested and never resolved, breaking the coordination loop.}{The team proceeds without an agreed decision; implementation assumptions diverge.}

\symptomexample{Unverifiable claims}{%
  A claim: max\_resolution is already added at specific line ranges.\par
  B constraint: I cannot verify your intermediate changes (separate branches).\par
  Result: the merge can be clean while one side's asserted change is silently absent.\par
}{The claim is specific but non-checkable; coordination lacks a verification mechanism (e.g., pasted signature, exact diff, or an agreed placeholder).}{False shared context about code state leads to incompatible downstream edits.}

\symptomexample{Broken commitment}{%
  A msg: I'll add BOTH parameters (fallback\_processor and max\_batch\_size) to the constructor signature, BOTH docstrings, and BOTH initializations.\par
  A msg: \checkmark Line 26: Added BOTH parameters (fallback\_processor and max\_batch\_size) to constructor signature.\par
  Observed after merge: constructor only has \texttt{fallback\_processor}, missing \texttt{max\_batch\_size}.\par
}{A makes a confident completion claim that is not corrected or verified by B; under partial observability, this creates false shared context.}{\texttt{TypeError: \_\_init\_\_() got an unexpected keyword argument max\_batch\_size}\@\ (tests fail).}

\symptomexample{Dependency access}{%
  Conversation evidence: \textbf{0 coordination messages} were sent (\texttt{total\_messages=0}), so neither agent disclosed import/initialization risks.\par
  Merge traceback excerpt: \texttt{src/PIL/Image.py:60} executes \texttt{from\ .\ import\ ImageDraw}. Then \texttt{src/PIL/ImageDraw.py:45} reads \texttt{Image.core}.\par
  (\texttt{PIL.Image} still initializing \(\Rightarrow\) circular import failure).\par
}{With no communication at all, the merged import graph is never discussed; the first shared integration check happens only at import time and fails deterministically.}{\texttt{AttributeError: partially initialized module PIL.Image has no attribute core (most likely due to a circular import)}.}

\symptomexample{Placeholder misuse}{%
  A msg: I'll add a clear comment marker: \texttt{[Conditional filters overlay insertion point]}.\par
  A msg: Please insert your logic immediately AFTER the marker\ldots without modifying lines above it.\par
  B msg: Given your marker plan, I didn't alter those methods\ldots I rely on \texttt{\_\_post\_init\_\_} to overlay filters.\par
}{The agreed integration point (insert-after-marker) is not used; B implements an alternative wiring path, so the merged decorator surface no longer matches the expected call pattern.}{\texttt{TypeError: prompt got an unexpected keyword argument conditional\_filters}.}

\symptomexample{Parameter flow}{%
  A msg: renamed \texttt{edit\_file} to \texttt{edit\_files} with multi-file command construction.\par
  B msg: I'm going to continue\ldots based on the current state I see (\texttt{edit\_file} method).\par
  B code shape: builds a shell command by interpolating \texttt{filename} into a quoted string, assuming it is a single string.\par
}{Ambiguity about a changing interface leaves one agent implementing against an outdated contract; after merge, a list flows into string-only formatting.}{\texttt{sed: can't read [\ldots]: No such file or directory} (list passed as a literal string).}

\symptomexample{Timing dependency}{%
  A msg: Processing Pipeline: load \(\to\) image.load \(\to\) EXIF correction (NEW) \(\to\) B crop (pending) \(\to\) mode conversion \(\to\) return.\par
  B msg: Applied AFTER EXIF correction (A) and BEFORE mode conversion\ldots Pipeline (after merge): load \(\to\) EXIF correction \(\to\) center-crop.\par
  Merge: CLEAN (0 textual conflicts); both declare No conflicts expected.\par
  Merged code excerpt: \texttt{image = image.crop(\ldots)}\par
  Merged code absence: no \texttt{ImageOps.exif\_transpose(\ldots)} call exists in the merged function.\par
}{They agree on the intended order, but fail to ensure the EXIF correction block is actually present at the agreed insertion point after merge.}{\texttt{assert (640, 480) == (480, 640)} (EXIF correction missing).}

\section{Case Study: Spatial vs.\ Semantic Coordination}\label{app:case_study}

Section~\ref{sec:role_of_communication} shows that communication reduces merge conflicts but does not improve task success. To understand why, we examine a representative failure in detail. This case illustrates the distinction between \emph{spatial coordination} (agreeing on which lines to edit) and \emph{semantic coordination} (agreeing on what values and behaviors to implement). We find that agents excel at the former but neglect the latter.

\subsection{Task Setup}

The task comes from the Jinja2 template engine. Jinja2 provides a \texttt{groupby} filter that groups items by an attribute. For example, \texttt{users|groupby("city")} groups a list of users by their city. The original function signature looks like this:

\begin{lstlisting}[language=Python,basicstyle=\small\ttfamily]
def sync_do_groupby(
    environment: "Environment",
    value: "t.Iterable[V]",
    attribute: t.Union[str, int],
    default: t.Optional[t.Any] = None,
) -> "t.List[t.Tuple[t.Any, t.List[V]]]":
\end{lstlisting}

Two agents must independently extend this filter with new parameters. Agent~1 adds a \texttt{case\_sensitive} parameter that controls whether grouping treats ``CA'' and ``ca'' as the same value. The intended default is \texttt{False}, matching the existing \texttt{sort} filter's case-insensitive behavior. Agent~2 adds a \texttt{reverse} parameter that controls whether groups appear in descending order.

The coordination challenge is that both parameters must be added to the same function signature. Git performs conflict detection at the \emph{hunk} level (contiguous blocks of changes), not individual lines. If both agents modify the signature independently, Git will detect a conflict because both changes touch the same hunk. This creates a genuine coordination requirement where agents cannot simply work in isolation.

\subsection{What Agents Communicated}

The agents exchanged 10 messages totaling over 3,000 words. Their communication focused almost entirely on spatial coordination: line numbers, file paths, and edit ranges.

\begin{tcolorbox}[symptomcard]
\small
\textbf{Agent 1:} I'll be modifying the \texttt{sync\_do\_groupby} and \texttt{do\_groupby} functions in \texttt{src/jinja2/filters.py} \textbf{lines 1161--1230}\ldots According to our coordination plan, both parameters will be added together to avoid merge conflicts.

\medskip

\textbf{Agent 2:} I'll wait for you to complete the implementation in \texttt{src/jinja2/filters.py} \textbf{lines 1161--1230}\ldots

\medskip

\textbf{Agent 1:} Your reverse insertion points remain:\\
\texttt{- sync: for \_, values in groupby(sorted(value, key=group\_expr), group\_expr):}\\
Please proceed to add \texttt{reverse: bool = False} immediately after \texttt{case\_sensitive}.
\end{tcolorbox}

This communication successfully identified the shared code region and established a protocol for who edits what. What the agents never discussed were the semantic details that matter for correctness. Agent~1 correctly states that \texttt{case\_sensitive} should default to \texttt{False} to match the \texttt{sort} filter, but this default value is never explicitly confirmed with Agent~2. Neither agent discusses whether the two parameters interact or whether their implementations are independent.

\subsection{What Went Wrong}

Because both agents anticipated the need to modify the same function signature, each independently added \emph{both} parameters to avoid a Git conflict. However, without semantic coordination, they made inconsistent choices.

Agent~1's patch added only the \texttt{case\_sensitive} parameter with the correct default:

\begin{lstlisting}[language=Python,basicstyle=\small\ttfamily]
def sync_do_groupby(
    environment: "Environment",
    value: "t.Iterable[V]",
    attribute: t.Union[str, int],
    default: t.Optional[t.Any] = None,
    case_sensitive: bool = False,  # Correct default
) -> "t.List[_GroupTuple]":
\end{lstlisting}

Agent~2's patch added \emph{both} parameters (to avoid merge conflicts), but reported the wrong value in communication:

\begin{tcolorbox}[evidencebox]
\small
\textbf{Agent 2's status message:}\\
``Signatures now are: \texttt{(environment, value, attribute, default=None, case\_sensitive=True)}''
\end{tcolorbox}

Agent~2 reported \texttt{case\_sensitive=True} as the default while the correct value is \texttt{False}. This discrepancy was never caught because the conversation focused entirely on \emph{where} edits would happen, not \emph{what values} would be used. Neither agent verified the other's actual implementation; they relied on status messages. The semantic meaning of the default (``should match the \texttt{sort} filter'') was mentioned by Agent~1 but never confirmed by Agent~2.

For reference, the gold (correct) patches show what each feature should look like. The gold patch for \texttt{case\_sensitive} adds:

\begin{lstlisting}[language=Python,basicstyle=\small\ttfamily]
    default: t.Optional[t.Any] = None,
    case_sensitive: bool = False,
) -> "t.List[_GroupTuple]":
\end{lstlisting}

And the gold patch for \texttt{reverse} adds:

\begin{lstlisting}[language=Python,basicstyle=\small\ttfamily]
    default: t.Optional[t.Any] = None,
    reverse: bool = False,
) -> "t.List[t.Tuple[t.Any, t.List[V]]]":
\end{lstlisting}

The correct merged signature would combine both:

\begin{lstlisting}[language=Python,basicstyle=\small\ttfamily]
def sync_do_groupby(
    environment: "Environment",
    value: "t.Iterable[V]",
    attribute: t.Union[str, int],
    default: t.Optional[t.Any] = None,
    case_sensitive: bool = False,
    reverse: bool = False,
) -> "t.List[_GroupTuple]":
\end{lstlisting}

\subsection{What Would Have Worked}

For this task to succeed, agents needed to coordinate on three levels. \emph{Spatial coordination} they achieved: ``I'm editing lines 1161--1230; please add your parameter after mine.'' \emph{Structural coordination} they partially achieved: ``Both parameters go in the signature; I'll add mine first.'' \emph{Semantic coordination} was missing entirely.

A single message could have prevented the failure:

\begin{tcolorbox}[evidencebox]
\small
\textbf{Missing coordination:}\\
``I'm implementing \texttt{case\_sensitive} with default value \texttt{False} (not \texttt{True}). This matches the \texttt{sort} filter's case-insensitive default. If you need to include this parameter in your patch, please use exactly \texttt{case\_sensitive: bool = False}.''
\end{tcolorbox}

\subsection{Implications}

This case study provides concrete evidence for the spatial-semantic gap discussed in Section~\ref{sec:role_of_communication}. Despite 10 messages and over 3,000 words of coordination, the agents never once discussed the actual default value that \texttt{case\_sensitive} should have. They successfully negotiated \emph{where} to edit but failed to negotiate \emph{what} to implement. A single clarifying message about the intended default value would have prevented the failure entirely.